%% file: main.tex
\documentclass{article}

\usepackage{microtype}
\usepackage{graphicx}
\usepackage{booktabs} 
\usepackage{subcaption}

\usepackage{hyperref}

\usepackage[accepted]{icml2020}

\usepackage{mycustom}
\loremipsumtrue 

\newcommand{\reals}{\mathbb{R}}

\newcommand{\nobs}{n_x}
\newcommand{\nstate}{n_s}
\newcommand{\ncontrol}{n_u}
 
\newcommand{\control}{u}
\newcommand{\Control}{U}
\newcommand{\sysnoise}{w} 
 
\newcommand{\state}{s} 
\newcommand{\obs}{x}
\newcommand{\latent}{z} 
\newcommand{\enc}{E}
\newcommand{\dyn}{F}
\newcommand{\dec}{D}

\newcommand{\Zspace}{\mathcal{Z}}
\newcommand{\Xspace}{\mathcal{X}}

\newcommand{\Uspace}{\mathcal{U}}
\newcommand{\tpo}{{t+1}}
\newcommand{\cons}{{\textrm{cons}}}
\newcommand{\curv}{{\textrm{curv}}}
\newcommand{\cpc}{{\textrm{cpc}}}

\newcommand{\mi}{{\textrm{MI}}}
\newcommand{\conse}{{\textrm{cons} + \eps}}
\newcommand{\cpce}{{\textrm{cpc} + \eps}}
\newcommand{\pred}{{\textrm{pred}}}
\newcommand{\bbP}{\mathbb{P}}
\newcommand{\ssj}{{(j)}}
\newcommand{\ssi}{{(i)}}
\DeclareMathOperator{\sign}{sign}
\newcommand{\sharedtitle}{
Predictive Coding for Locally-Linear Control
}
\icmltitlerunning{\sharedtitle}

\begin{document}

\twocolumn[
\icmltitle{\sharedtitle}



\icmlsetsymbol{equal}{*}

\begin{icmlauthorlist}
\icmlauthor{Rui Shu}{equal,stanford}
\icmlauthor{Tung Nguyen}{equal,vinai}
\icmlauthor{Yinlam Chow}{google}
\icmlauthor{Tuan Pham}{vinai}
\icmlauthor{Khoat Than}{vinai}
\icmlauthor{Mohammad Ghavamzadeh}{facebook}
\icmlauthor{Stefano Ermon}{stanford}
\icmlauthor{Hung Bui}{vinai}
\end{icmlauthorlist}

\icmlaffiliation{stanford}
{Stanford University}
\icmlaffiliation{vinai}
{VinAI}
\icmlaffiliation{google}
{Google Research}
\icmlaffiliation{facebook}
{Facebook AI Research}

\icmlcorrespondingauthor{Rui Shu}{ruishu@stanford.edu}
\icmlcorrespondingauthor{Tung Nguyen}{v.tungnd13@vinai.io}

\icmlkeywords{Machine Learning, ICML}

\vskip 0.3in
]



\printAffiliationsAndNotice{\icmlEqualContribution} 

\begin{abstract}
High-dimensional observations and unknown dynamics are major challenges when applying optimal control to many real-world decision making tasks.
The Learning Controllable Embedding (LCE) framework addresses these challenges by embedding the observations into a lower dimensional latent space, estimating the latent dynamics, and then performing control directly in the latent space.
To ensure the learned latent dynamics are predictive of next-observations, all existing LCE approaches decode back into the observation space and explicitly perform next-observation prediction---a challenging high-dimensional task that furthermore introduces a large number of nuisance parameters (i.e., the decoder) which are discarded during control.
In this paper, we propose a novel \emph{information-theoretic} LCE approach and show theoretically that explicit next-observation prediction can be replaced with predictive coding. We then use predictive coding to develop a \emph{decoder-free} LCE model whose latent dynamics are amenable to locally-linear control. 
Extensive experiments on benchmark tasks show that our model reliably learns a controllable latent space that leads to superior performance when compared with state-of-the-art LCE baselines.
\end{abstract}

\ifdefined\isaccepted
\vspace{-0.5cm}
\fi
\section{Introduction}

With the rapid growth of systems equipped with powerful sensing devices, it is important to develop algorithms that are capable of controlling systems from high-dimensional raw sensory inputs (e.g., pixel input). However, scaling stochastic optimal control and reinforcement learning (RL) methods to high-dimensional unknown environments remains an open challenge. To tackle this problem, a common approach is to employ various heuristics to embed the high-dimensional observations into a lower-dimensional latent space \cite{finn2016deep,kurutach2018learning,kaiser2019model}. The class of Learning Controllable Embedding (LCE) algorithms  \cite{watter2015embed,banijamali2017robust,hafner2018learning,zhang2019solar,levine2020prediction} further supplies the latent space with a latent dynamics model to enable planning directly in latent space.


Our present work focuses on this class of LCE algorithms and takes a critical look at the prevailing heuristic used to learn the controllable latent space: next-observation prediction. To ensure that the learned embedding and latent dynamics are predictive of future observations, existing LCE algorithms introduce a decoder during training and explicitly perform next-observation prediction by decoding the predicted latent states back into the observation space. Despite its empirical success \cite{watter2015embed,banijamali2017robust,zhang2019solar,levine2020prediction}, this approach suffers from two critical drawbacks that motivate the search for better alternatives: 
(i) it requires the model to handle the challenging task of high-dimensional prediction; 
(ii) it does so in a parameter-inefficient manner---requiring the use of a decoder that is discarded during control. 

To address these concerns, we propose a novel information-theoretic LCE approach for learning a controllable latent space. Our contributions are as follows.
\begin{enumerate}
    \item We characterize the quality of the learned embedding through the lens of \emph{predictive suboptimality} and show that predictive coding \cite{oord2018representation} is sufficient for minimizing predictive suboptimality.
    \item Based on predictive coding, we propose a simpler and parameter-efficient model that jointly learns a controllable latent space and latent dynamics specifically amenable to locally-linear controllers.
    \item We conduct detailed analyses and empirically characterize how model ablation impacts the learned latent space and control performance.
    \item Finally, we show that our method out-performs state-of-the-art LCE algorithms on several benchmark tasks, demonstrating predictive coding as a superior alternative to next-observation prediction when learning controllable embeddings.
\end{enumerate}

\section{Background}\label{sec:background}
We are interested in controlling non-linear dynamical systems of the form $\state_{t+1} = f_{\mathcal S}(\state_t, \control_t)+\sysnoise$, over the horizon $T$. In this definition, $\state_t \in \mathcal S\subseteq\reals^{\nstate}$ and $\control_t \in \mathcal U\subseteq\reals^{\ncontrol}$ are the state and action of the system at time step $t\in \{0,\ldots,T-1\}$, $\sysnoise$ is the Gaussian system noise, and $f_{\mathcal S}$ is the smooth non-linear system dynamics. 
We are particularly interested in the scenario in which we only have access to the high-dimensional observation $\obs_t \in \mathcal X\subseteq\reals^{\nobs}$ of each state $s_t$ ($\nobs \gg \nstate$). This scenario has application in many real-world problems, such as visual-servoing \cite{espiau1992new}, in which we only observe high-dimensional images of the environment and not its underlying state. We further assume that the high-dimensional observations $x$ have been selected such that for any arbitrary control sequence $\Control=\{\control_t\}_{t=0}^{T-1}$, the observation sequence $\{\obs_t\}_{t=0}^{T}$ is generated by a stationary Markov process, i.e.,~$x_{t+1}\sim p(\cdot|x_t,u_t),\;\forall t\in\{0,\ldots,T-1\}$.\footnote{One method to enable this Markovian assumption is by buffering observations~\citep{mnih2013playing} for a number of time steps.} 


A common approach to control the non-linear dynamical system described above is to solve the following stochastic optimal control (SOC) problem \cite{shapiro2009lectures} that minimizes the expected cumulative cost 
\begin{equation}
\tag{SOC1}
\label{problem:soc1}
\min_{\Control} \; L(\Control, p, c,\obs_0)
:=\Expect \Big[\sum_{t=0}^{T-1}c(\obs_t,\control_t)\mid p,\obs_0\Big],
\end{equation}
where $c:\mathcal X\times\mathcal U\rightarrow\reals_{\geq 0}$ is the immediate cost function and $\obs_0$ is the observation at the initial state $s_0$. 
Throughout the paper, we assume that all immediate cost functions are bounded by $c_{\max}>0$ and Lipschitz with constant $c_{\text{lip}}>0$. One form of the immediate cost function that is particularly common in goal tracking problems is $c(x,u)=\|x-\obs_{\text{goal}}\|^2$, where $\obs_{\text{goal}}$ is the observation at the goal state. 


The application of SOC to high-dimensional environments, however, faces several challenges. Since the observations $x$ are high-dimensional and the dynamics in the observation space $p(\cdot|x_t,u_t)$ is unknown, solving~\eqref{problem:soc1} is often intractable as it requires solving two difficult problems: high-dimensional dynamics estimation and high-dimensional optimal control. 
To address these issues, the Learning Controllable Embedding (LCE) framework proposes to learn a low-dimensional latent (embedding) space $\mathcal Z\subseteq\reals^{n_z}$ ($n_z \ll n_x$) and a latent state dynamics, and then perform optimal control in the latent space instead. This framework includes algorithms such as E2C~\citep{watter2015embed}, RCE~\citep{banijamali2017robust}, SOLAR~\citep{zhang2019solar}, and PCC~\citep{levine2020prediction}. By learning a stochastic encoder $\enc:\mathcal X\rightarrow\mathbb{P}(\mathcal Z)$ and latent dynamics $\dyn:\Zspace\times \Uspace \rightarrow \mathbb P(\Zspace)$, LCE algorithms defines a new SOC in the latent space,
\begin{align}
\label{problem:soc2}
\tag{SOC2}
\min_{\Control} \;  \Expect\Big[L(\Control,\dyn,\overline c,z_0)\mid \enc(x_0)\Big],
\end{align}
where $z_0$ is sampled from the distribution $E(x_0)$, i.e.,~$z_0 \sim E(z_0 \giv x_0 )$, and $\bar c:\mathcal Z\times\mathcal U\rightarrow\reals_{\geq 0}$ is the latent cost function. By solving the much lower-dimensional \eqref{problem:soc2}, the resulting optimal control $U^*_2$ is then applied as a feasible solution to~\eqref{problem:soc1} and incurs a suboptimality that depends on the choice of the encoder $E$ and latent dynamics $F$.\footnote{This suboptimality also depends on $\bar c$, but we assume $\bar c$ to be simple, e.g.,~$\bar c(z, u) = \|z - z_\textrm{goal}\|^2$, where $z_\textrm{goal}=E(x_\textrm{goal})$. $E$ thus subsumes the responsibility of defining a latent space that is compatible with $\bar c$. See Appendix \ref{proof:soc1soc1-e} for further justification.} 

Although \citet{levine2020prediction} provided an initial theoretical characterization of this SOC suboptimality, the selection of $E$ and $F$ ultimately remains heuristically-driven in all previous works. These heuristics vary across different studies \cite{levine2020prediction,banijamali2017robust,watter2015embed,zhang2019solar,hafner2018learning}, but the primary approach employed by the existing LCE algorithms is explicit next-observation prediction. By introducing a decoder $D: \Z \to \bbP(\X)$, the composition $D \circ F \circ E$ is cast as an action-conditional latent variable model; the advances in latent variable modeling \cite{kingma2013auto,rezende2014stochastic,burda2015importance, johnson2016composing,sohn2015learning} are then leveraged to train $E$, $F$, and $D$ to perform explicit next-observation prediction by maximizing a lower bound on the log-likelihood $
    \ln \int D(x_\tpo \giv z_\tpo) F(z_\tpo \giv z_t, u_t) E(z_t \giv x_t) \d {z_{t:\tpo}}
$, over the dataset whose trajectories are drawn from $p(x_t, u_t, x_\tpo)$. 

Next-observation prediction offers a natural way to learn a non-degenerate choice of $E$ and $F$, and enjoys the merit of being empirically successful. However, it requires the introduction of a decoder $D$ as nuisance parameter that only serves the auxiliary role of training the encoder $E$ and latent dynamics $F$. The focus of our paper is whether $E$ and $F$ can be successfully selected via a decoder-free heuristic. 




\input{figures/model}

\section{Information-Theoretic LCE}\label{sec:itlce}
Existing methods instantiate \eqref{problem:soc2} by learning the encoder $E$ and latent dynamics model $F$ in conjunction with an auxiliary decoder $D$ to explicitly perform next-observation prediction. The auxiliary decoder ensures that the learned representation \emph{can} be used for next-observation prediction, and is discarded after the encoder and latent dynamics model are learned. Not only is this a parameter-inefficient procedure for learning $(E, F)$, this approach also learns $(E, F)$ by explicitly performing the challenging high-dimensional next-observation prediction. In this section, we propose an information-theoretic approach that can learn $(E, F)$ without decoding and next-observation prediction. 

\subsection{Predictive Suboptimality of a Representation}
\label{subsec:pred_subopt}

Our approach exploits the observation that the sole purpose of the decoder is to ensure that the learned representation is good for next-observation prediction. In other words, the decoder is used to characterize the suboptimality of next-observation prediction when the prediction model is forced to rely on the learned representation. We refer to this concept as \emph{predictive suboptimality} of the learned representation and formally define it as follows.

\begin{definition}
\label{def:pred_subopt}
Let $p(x_\tpo, x_t, u_t)$ denote the data distribution. Given an encoder $E: \X \to \Z$, \footnote{For simplicity, we assume that the encoder $E$ considered here is deterministic.} let $q(x_\tpo \giv x_t, u_t)$ denote the prediction model
\begin{align*}
    q(x_\tpo \giv x_t, u_t) \propto \psi_1(x_\tpo) \psi_2(E(x_\tpo), E(x_t), u_t),
\end{align*}
where $\psi_1$ and $\psi_2$ are expressive non-negative functions. We define the predictive suboptimality $\ell^*_\pred(E)$ of a representation induced by $E$ as the best-case prediction loss
\begin{align*}
    \min_q \Expect_{p(x_\tpo, x_t, u_t)} D_{\textrm{KL}}\left[{p(x_\tpo \giv x_t, u_t)}||{q(x_\tpo \giv x_t, u_t)}\right].\label{eq:prediction}
\end{align*}
\end{definition}
Importantly, the function $\psi_2$ should measure the compatibility of the triplet $(x_\tpo, x_t, u_t)$---but is only allowed to do so via the representations $E(x_\tpo)$ and $E(x_t)$. Thus, the behavior of the representation bottleneck plays a critical role in modulating the expressivity of the model $q$. If $E$ is invertible, then $q$ is a powerful prediction model; if $E$ is a constant, then $q$ can do no better than marginal density estimation of $p(x_\tpo)$. 

While it is possible to minimize the predictive suboptimality of $E$ by introducing the latent dynamics model $F$ and decoder $D$, and then performing next-observation prediction via $D\circ F \circ E$, our key insight is that predictive suboptimality can be bounded by the following mutual information gap (see Appendix \ref{proof:pred_subopt} for proof).
\begin{lemma}
\label{lem:pred_subopt}
Let $X_\tpo$, $X_t$, and $U_t$ be the random variables associated with the data distribution $p(x_\tpo, x_t, u_t)$. The predictive suboptimality $\ell^*_\pred(E)$ is upper bounded by the mutual information gap
\begin{align*}
    I(X_\tpo \scolon X_t, U_t) - I(E(X_\tpo) \scolon E(X_t), U_t).
\end{align*}
\end{lemma}
Since $I(X_\tpo \scolon X_t, U_t)$ is a constant and upper bounds $I(E(X_\tpo) \scolon E(X_t), U_t)$ by the data processing inequality, this means we can minimize the predictive suboptimality of $E$ simply by maximizing the mutual information between the future latent state $E(X_\tpo)$ and the current latent state and action pair $(E(X_t), U_t)$---a form of predictive coding. We denote this mutual information $\ell_\mi(E)$ as a function of $E$. To maximize this quantity, we can then leverage the recent advances in variational mutual information approximation \cite{oord2018representation,poole2019variational,belghazi2018mutual,nguyen2010estimating,hjelm2018learning} to train the encoder in a decoder-free fashion.

\subsection{Consistency in Prediction of the Next Latent State}\label{subsec:consistency}

A notable consequence of introducing the encoder $E$ is that it can be paired with a latent cost function $\bar c$ to define an alternative cost function in the observation space,
\begin{align*}
    c_E(x, u) := \Expect\Big[\bar c(z, u) \mid \enc(x)\Big],
\end{align*}
where $z$ is sampled from $E(x)$.\footnote{In \cref{subsec:consistency,subsec:ilqr_latent}
, we consider the general case of the stochastic encoder in order to extend the analysis in \cite{levine2020prediction}. This analysis readily carries over to the limiting case when $E$ becomes deterministic.}
This is particularly useful for high-dimensional SOC problems, where it is difficult to prescribe a meaningful cost function \emph{a priori} in the observation space. For example, for goal tracking problems using visuosensory inputs, prescribing the cost function to be $c(x,u)=\|x-\obs_{\text{goal}}\|^2$ suffers from the uninformative nature of the $2$-norm in high-dimensional pixel space \cite{beyer1999nearest}. In the absence of a prescribed $c$, a natural proxy for the unknown cost function is to replace it with $c_E$ and consider the new SOC problem,
\begin{equation}
\tag{SOC1-E}
\label{problem:soc1-e}
\min_{\Control} \; L(\Control, p, c_E,\obs_0).
\end{equation}
Assuming \eqref{problem:soc1-e} to be the de facto SOC problem of interest, we wish to learn an $F$ such that the optimal control $U^*_2$ in \eqref{problem:soc2} approximately solves \eqref{problem:soc1-e}. One such consideration for the latent dynamics model would be to set $F$ as the true latent dynamics induced by $(p, E)$, and we refer to such $F$ as the one that is \emph{consistent} with $(p, E)$.

Our main contribution in this section is to justify---from a control perspective---why selecting a consistent $F$ with respect to $(p, E)$ minimizes the suboptimality incurred from using \eqref{problem:soc2} as an approximation to \eqref{problem:soc1-e}. The following lemma (see Appendix \ref{proof:consistency} for proof) provides the suboptimality performance gap between the solutions of~\eqref{problem:soc2} and~\eqref{problem:soc1-e}.
\begin{lemma}
\label{lem:consistency}
For any given encoder $\enc$ and latent dynamics $\dyn$, let $U^*_{\text{1-E}}$ be the solution to~\eqref{problem:soc1-e} and $U^*_{\text{2}}$ be a solution to~\eqref{problem:soc2}. Then, we have the following performance bound between the costs of the control signals $U^*_{\text{1-E}}$ and $U^*_{\text{2}}$: 
\begin{equation}\label{eq:consistency}
L(\Control^*_{\text{1-E}}, p, c_E,\obs_0) \geq L(\Control^*_{\text{2}}, p, c_E,\obs_0)- 2\lambda_{\text{C}}\cdot \sqrt{2R_{\text{C}}(\enc, \dyn)},
\end{equation}
where
$
R_{\text{C}}(\enc,\dyn) =  
\mathbb E_{p(x_\tpo, x_t, u_t)}
[D_{\text{KL}}(\enc(z_\tpo|x_\tpo)||(F\circ E)(z_\tpo|x_t,u_t))]$ and $\lambda_{\text{C}} = T^2c_{\text{max}}\overline U$.
\end{lemma}
In Eq.~\ref{eq:consistency}, the expectation is over the state-action stationary distribution of the policy used to generate the training samples (uniformly random policy in this work), and $\overline U$ is the \emph{Lebesgue measure} of $\mathcal U$.\footnote{In the case when sampling policy is non-uniform and has no measure-zero set, $1/\overline U$ is its minimum measure.} Moreover, $E(z_\tpo|x_\tpo)$ and $\big(F\circ E\big)(z_\tpo|x_t,u_t) = \int F(z_\tpo|z_t,u_t)E(z_t|x_t) \d z_t$ are the probability over the next latent state $z_\tpo$. 
Based on \Cref{fig:model}, we therefore interpret $R_{\text{C}}(\enc,\dyn)$ as the measure of discrepancy between the dynamics $x_t \to x_\tpo \to z_\tpo$ induced by $(p, E)$ versus the latent dynamics model $x_t \to z_t \to z_\tpo$ induced by $(E, F)$. which we term the \emph{consistency} regularizer. We note that while our resulting bound is similar to Lemma~2 in~\citet{levine2020prediction}, there are two key differences. First, our analysis makes explicit the assumption that the cost function $c$ is not prescribed and thus replaced in practice with the proxy cost function $c_E$ based on the heuristically-learned encoder. Second, by making this assumption explicit, our bound is based on samples from the environment dynamics $p$ instead of the next-observation prediction model dynamics $\hat p$ as required in~\citet{levine2020prediction}.

By restricting the stochastic encoder $E$ to be a distribution with fixed entropy (e.g.,~by fixing the variance if $E$ is conditional Gaussian), the minimization of the consistency regularizer corresponds to maximizing the log-likelihood of $F$ for predicting $z_\tpo$, given $(z_t, u_t)$, under the dynamics induced by $(p,E)$. This correspondence holds even in the limiting case of $E$ being deterministic (e.g.,~fixing the variance to an arbitrarily small value). In other words, for~\eqref{problem:soc2} to approximate~\eqref{problem:soc1-e} well, we select $F$ to be a good predictor of the true latent dynamics.

\subsection{Suboptimality in Locally-Linear Control} 
\label{subsec:ilqr_latent}
In Section~\ref{subsec:consistency}, we derived the suboptimality of using \eqref{problem:soc2} as a surrogate control objective for \eqref{problem:soc1-e}, and showed that the suboptimality depends on the consistency of latent dynamics model $F$ with respect to the true latent dynamics induced by $(p, E)$. 

We now shift our attention to the optimization of \eqref{problem:soc2} itself. Similar to previous works \cite{watter2015embed,banijamali2017robust,zhang2019solar,levine2020prediction}, we shall specifically consider the class of locally-linear control (LLC) algorithms, e.g.,~iLQR~\citep{li2004iterative}, for solving~\eqref{problem:soc2}.
The main idea in LLC algorithms is to compute an optimal action sequence by linearizing the dynamics around some nominal trajectory. This procedure implicitly assumes that the latent dynamics $\dyn$ has low curvature, so that local linearization via first-order Taylor expansion yields to a good linear approximation over a sufficiently large radius. As a result, the \emph{curvature} of $F$ will play an important role in the optimizability of \eqref{problem:soc2} via LLC algorithms. 

\citet{levine2020prediction} analyzed the suboptimality incurred from applying LLC algorithms to \eqref{problem:soc2} as a function of the curvature of $F$. For self-containedness, we summarize their analysis as follows. We shall assume $F$ to be a conditional Gaussian model with a mean prediction function $f_\Z(z,u)$. The curvature of $f_\Z$ can then be measured via
\begin{align*}
R_{\text{LLC}}(\dyn) = \mathbb E_{\obs,\control,\eta}\big[
\|
f_{\mathcal Z}(\latent+\eta_\latent,\control+\eta_\control) 
- f_{\mathcal Z}(\latent, \control)\\
- (\nabla_{\latent}f_{\mathcal Z}(\latent, \control)\cdot\eta_\latent + \nabla_{\control}f_{\mathcal Z}(\latent, \control)\cdot\eta_\control) 
\|_{2}^{2} 
\mid \enc\big].
\end{align*}
where  $\eta=(\eta_z,\eta_u)^\top\sim\mathcal N(0,\delta^2 I)$, $\delta>0$ is a tunable parameter that characterizes the radius of latent state-action space in which the latent dynamics model should have low curvature. 
Let $U^*_{\text{LLC}}$ be a LLC solution to~\eqref{problem:soc2}. Suppose the nominal latent state-action trajectory $\{(\latent_t,\control_t)\}_{t=0}^{T-1}$ satisfies the condition: $(\latent_t,\control_t)\sim\mathcal N((\latent^*_{2,t},\control^*_{2,t}),\delta^2I)$, where $\{(\latent^*_{2,t},\control^*_{2,t})\}_{t=0}^{T-1}$ is the optimal trajectory of \eqref{problem:soc2}.
Using Eq.~29 of \citet{levine2020prediction}, one can show that with probability $1-\eta$, the LLC solution of~\eqref{problem:soc2} has the following suboptimality performance gap when compared with the optimal cost of this problem using the solution $U^*_2$,
 \begin{align*}\label{eq:curvature}
 L(U^*_{\text{2}},F, \bar c,z_0)\geq L(\Control^*_{\text{LLC}}, F, \bar c,z_0) - 2\lambda_{\text{LLC}}\cdot\sqrt{R_{\text{LLC}}(F)},
 \end{align*}
 where 
 \[
 \lambda_{\text{LLC}} = T^2c_{\max}c_{\text{lip}}  (1+\sqrt{2\log(2T/\eta)})\sqrt{\overline U\overline X}/2,
 \]
 and $\overline{X}$ is the Lebesgue measure with respect to~$\Xspace$. We therefore additionally constrain $F$ to have low curvature so that it is amenable to the application of LLC algorithms.


\section{Predictive Coding, Consistency, Curvature}
Based on the analysis in \cref{sec:itlce}, we identify three desiderata for guiding the selection of the encoder $E$ and latent dynamics model $F$. We summarize them as follows: 
(i) {\em predictive coding} minimizes the predictive suboptimality of the encoder $E$;
(ii) {\em consistency} of the latent dynamics model $F$ with respective to $(p, E)$ enables planning directly in the latent space; and
(iii) {\em low-curvature} enables planning in latent space specifically using locally-linear controllers.
We refer to these heuristics collectively as Predictive Coding-Consistency-Curvature (PC3). PC3 can be thought of as an information-theoretic extension of the Prediction-Consistency-Curvature (PCC) framework described by \citet{levine2020prediction}---differing primarily in the replacement of explicit next-observation prediction with predictive coding in the latent space.

In this section, we highlight some of the key design choices involved when instantiating PC3 in practice. In particular, we shall show how to leverage the CPC variational mutual information bound in a parameter-efficient manner and how to enforce the consistency of $F$ with respect to $(p, E)$ without destabilizing training.

\subsection{Enforcing Predictive Codes}
To estimate the mutual information $\ell_\mi(E)$, we employ contrastive predictive coding (CPC) proposed by \citet{oord2018representation}. We perform CPC by introducing a critic $f: \Z \times \Z \times \U \to \R$ to construct the lower bound
\begin{align}
    &I(E(X_\tpo) \scolon E(X_t), U_t)\label{eq:cpc} \\
    &\hspace{0.5cm}\ge 
    \Expect 
    \frac{1}{K} \sum_i
    \ln \frac
        {\exp f(E(x_\tpo^\ssi), E(x_t^\ssi), u_t^\ssi)}
        {\frac{1}{K}\sum_j \exp f(E(x_\tpo^\ssi),  E(x_t^\ssj), u_t^\ssj)},\nonumber
\end{align}
where the expectation is over $K$ i.i.d. samples of $(x_\tpo, x_t, u_t)$. Notice that the current latent state-action pair $(E(x_t), u_t)$ is specifically designated as the source of negative samples and used for the contrastive prediction of the next latent state $E(x_\tpo)$. We then tie the critic $f$ to our latent dynamics model $F$,
\begin{align*}
    \exp f(z_\tpo, z_t, u_t) := F(z_\tpo \giv z_t, u_t).
\end{align*}
This particular design of the critic has two desirable properties. First, it exploits parameter-sharing to circumvent the instantiation of an auxiliary critic $f$. Second, it takes advantage of the property that an optimal critic for the lower bound in \cref{eq:cpc} is the true latent dynamics \cite{poole2019variational,ma2018noise}---which we wish $F$ to approximate. The resulting CPC objective is thus
\begin{align*}
    \Expect 
    \frac{1}{K} \sum_i
    \ln \frac
        {F(E(x_\tpo^\ssi) \giv E(x_t^\ssi), u_t^\ssi)}
        {\frac{1}{K}\sum_j F(E(x_\tpo^\ssi) \giv  E(x_t^\ssj), u_t^\ssj)},
\end{align*}
which we denote as $\ell_\cpc(E, F)$.





\subsection{Enforcing Consistency}\label{subsec:enforce_consistency}
Since the true latent dynamics is an optimal critic for the CPC bound, it is tempting to believe that optimizing $(E, F)$ to maximize $\ell_\cpc(E, F)$ should be sufficient to encourage the learning of a latent dynamics model $F$ that is consistent with the true latent dynamics induced by $(p, E)$. 

In this section, we show that it is easy to construct a simple counterexample illustrating the non-uniqueness of the true latent dynamics as an optimal critic---and that $F$ may learn to be arbitrarily \emph{inconsistent} with $(p, E)$ while still maximizing $\ell_\cpc(E, F)$ under a fixed choice of $E$. Our simple counterexample proceeds as follows: let $E$ be the identity function, let $\X = \U = \R$, and let $p(x_\tpo, x_t, u_t)$ be a uniform distribution over the tuples $(1, 1, 1)$ and $(-1, -1, -1)$. Let $F(z_\tpo \giv z_t, u_t)$ be a conditional Gaussian distribution with learnable variance $\sigma^2 > 0$ and mean function
\begin{align*}
    \mu(z_t, u_t) = \sign(z_t) \cdot \eta,
\end{align*}
where $\eta > 0$ is a learnable parameter. By symmetry, the bound $\ell_\cpc(E, F)$ where $K = 2$ becomes
\begin{align*}
    \ln \frac{\exp((\eta - 1)^2 / \sigma^2)}{ \exp((\eta - 1)^2 / \sigma^2) + \exp((\eta + 1)^2/ \sigma^2)} + \ln 2.
\end{align*}
In the denominator, the first term arises from the positive sample (e.g., $(1, 1, 1)$) whereas the second term arises from the negative sample (e.g., $(1, -1, -1)$). One way to maximize this bound would be to set $\eta = 1$ and let $\sigma \to 0$. Correspondingly, $F$ would approach the true latent dynamics and precisely predict how $(z_t, u_t)$ transitions to $z_\tpo$. However, an alternative procedure for maximizing this bound is to fix $\sigma$ to any positive constant and let $\eta \to \infty$. In this scenario, $F$ becomes an arbitrarily poor predictor of the underlying latent dynamics. 

This counterexample highlights a simple but important characteristic of the CPC bound. In contrast to direct maximum likelihood training of $F(z_\tpo \giv z_t, u_t)$ using samples of $(z_\tpo, z_t, u_t)$ from the true latent dynamics, the contrastive predictive training of the latent dynamics model simply ensures that $F(z_\tpo \giv z_t, u_t)$ assigns a \emph{relatively} much higher value to the positive samples than to the negative samples. The fact that the CPC bound may be maximized without learning a consistent dynamics model $F$ may be why previous work by \citet{nachum2018near} using CPC for representation learning in model-free RL chose not to perform model-based latent space control despite also learning an $F$ as a variational artifact from their CPC bound.

Since our goal is to use $F$ in \eqref{problem:soc2} for optimal control, it is critical that we ensure the latent dynamics model $F$ indeed approximates the true latent dynamics. We therefore additionally train $F$ via the maximum likelihood objective
\begin{align*}
    \ell_\cons(E, F) = \Expect_{p(x_\tpo, x_t, u_t)} \ln F(E(x_\tpo) \giv E(x_t),  u_t).
\end{align*}
However, naively optimizing $(E, F)$ to maximize both $\ell_\cpc$ and $\ell_\cons$ is unstable; whereas $\ell_\cpc$ is geometry-invariant,  $\ell_\cons$ is sensitive to non-volume preserving transformations of the latent space \cite{rezende2015variational,dinh2016density} and can increase arbitrarily simply by collapsing the latent space. To resolve this issue, we add Gaussian noise $\eps \sim \Normal(0, \sigma^2 I)$ with fixed variance to the next-state encoding $E(x_\tpo)$. Doing so yields the noise-perturbed objectives $\ell_\cpce$ and $\ell_\conse$. The introduction of noise has two notable effects. First, it imposes an upper bound on the achievable log-likelihood
\begin{align*}
    \ell_\conse(E, F) \le -\frac{n_z}{2} \ln 2\pi e \sigma^2
\end{align*}
based on the entropy of the Gaussian noise. Second, $\ell_\cpce$ is now a lower bound to the mutual information between $(E(X_t), U_t)$ and the noise-perturbed $E(X_\tpo) + \mathcal{E}$,
\begin{align*}
    &I(E(X_\tpo) + \mathcal{E} \scolon E(X_t), U_t) \\
    &\ge 
    \Expect 
    \frac{1}{K} \sum_i
    \ln \frac
        {F(E(x_\tpo^\ssi)+ \eps^\ssi \giv E(x_t^\ssi), u_t^\ssi)}
        {\frac{1}{K}\sum_j F(E(x_\tpo^\ssi) + \eps^\ssi \giv  E(x_t^\ssj), u_t^\ssj)}.\nonumber
\end{align*}
Since the noise variance $\sigma^2$ is fixed, $\ell_\cpce$ can only be maximized by expanding the latent space. By tuning the noise variance $\sigma^2$ as a hyperparameter, we can balance the latent space retraction encouraged by $\ell_\conse$ with the latent space expansion encouraged by $\ell_\cpce$ and thus stabilize the learning of the latent space. For notational simplicity, we shall treat all subsequent mentions of $\ell_\cpc$ and $\ell_\cons$ to mean their respective noise-perturbed variants, except in the specific ablation conditions where noise is explicitly removed (e.g., the ``w/o $\eps$'' condition in our experiments).

\subsection{Enforcing Low Curvature}
We measure the curvature of $F$ by computing the first-order Taylor expansion error incurred when evaluating at $\bar{z}=z+\eta_z$ and $\bar{u}=u+\eta_u$,
\[
\begin{split}
   &\ell_\curv(F)=
\mathbb E_{\eta\sim\mathcal N(0,\delta I)}[\| f_{\mathcal Z}(\bar{z},\bar{u})  - (\nabla_{z}f_{\mathcal Z}(\bar{z}, \bar{u})\eta_z \\
&\qquad\qquad+ \nabla_{u}f_{\mathcal Z}(\bar{z}, \bar{u})\eta_u) - f_{\mathcal Z} (z, u)\|_{2}^{2}].
\end{split}
\]
\citet{levine2020prediction} further proposes an amortized version of this objective to accelerate training when the latent dimensionality $n_z$ is large. However, since $n_z$ is relatively small in our benchmark tasks, our initial experimentation suggests amortization to have little wall-clock time impact on these tasks. Our overall objective is thus
\begin{align*}
    \max_{E, F} \lambda_1 \ell_\cpc(E, F) + \lambda_2 \ell_\cons(E, F) - \lambda_3 \ell_\curv(F),
\end{align*}
which maximizes the CPC bound and consistency, while minimizing curvature. 





\input{figures/map_ablation}
\input{tables/ablation}
\input{tables/latent_prediction}

\section{Experiments}
\input{figures/map_compare}
\input{tables/control_compare}

In this section, we report a thorough ablation study on various components of PC3, as well as compare the performance of our proposed model\footnote{
\ifdefined\isaccepted
\href{https://github.com/VinAIResearch/PC3-pytorch}{https://github.com/VinAIResearch/PC3-pytorch}
\else
Implementation will be available at code submission deadline.
\fi
} with two state-of-the-art LCE baselines: PCC~\cite{levine2020prediction} and SOLAR~\cite{zhang2019solar}.\footnote{E2C and RCE, two closely related baselines, are not included, since they are often inferior to PCC \cite{levine2020prediction}.} The experiments are based on four image-based control benchmark domains: Planar System, Inverted Pendulum,\footnote{Pendulum has two separate tasks: Balance and Swing Up} Cartpole, and 3-Link Manipulator.

\textbf{Data generation procedure:} In PCC and PC3, each sample is a triplet $(x_{t},u_{t},x_{t+1})$, in which we (1) sample uniformly an underlying state $s_{t}$ and generate its corresponding observation $x_{t}$, (2) sample uniformly an action $u_{t}$, and (3) obtain the next state $s_{t+1}$ from the true dynamics and generate the corresponding observation $x_{t+1}$. In SOLAR, each training sample is an episode $\{x_1,u_1,x_2,\dots,x_T,u_T,x_{T+1}\}$, where $T$ is the control horizon. We sample uniformly $T$ actions from the action space, apply the dynamics $T$ times from the initial state, and generate $T$ corresponding observations.

\textbf{Evaluation metric:} We evaluate PC3 and the baselines in terms of control performance. For PC3 and PCC, we apply iLQR algorithm in the latent space with a quadratic cost, $c(z_t,u_t) = (z_t-z_{\text{goal}})^\top Q (z_t-z_{\text{goal}}) + u_t^\top R u_t$, where $z_t$ and $z_{\text{goal}}$ are the encoded vectors of the current and goal observation, and $Q = \alpha \cdot I_{n_z}$, $R = \beta \cdot I_{n_u}$. For SOLAR, we use their original local-inference-and-control algorithm.\footnote{https://github.com/sharadmv/parasol} We report the percentage of time spent in the goal region in the underlying system~\cite{levine2020prediction}.

\subsection{Ablation Study}\label{subsec:ablation}

We characterize PC3 by ablating $\ell_\cons$, $\ell_\curv$, and the noise $\eps$ added to $z_\tpo$. For each setting, we report the latent map size,\footnote{We add the loss $||\frac{1}{N}\sum_{i=1}^N z_i||_2^2$ to center the latent map at the origin, then report $\frac{1}{N} \sum_{i=1}^N ||z_i||_2^2$ as the latent map size.} $\ell_\cpc$, $\ell_\cons$, $\ell_\curv$, and the control performance. These statistics are averaged over $10$ different models. All settings are run on Pendulum (Balance and Swing Up).

\textit{Consistency:} In Table~\ref{table:ablation}, we can see that when $\ell_\cons$ is omitted, the control performance drops. As discussed in \cref{subsec:consistency}, the latent dynamics model $F$ performs poorly when not explicitly optimized for consistency. This is further demonstrated in Table \ref{table:latent_prediction}, where we take a pretrained PC3 model, freeze the encoder, and retrain $F$ to maximize either $\ell_\cpc$ or $\ell_\cons$. Despite both retrained models achieving similar $\ell_\cpc$ scores, it is easy to see that training via $\ell_\cpc$ results in much worse latent dynamics in terms of prediction.

\textit{Noise:} The control performance also decreases when we do not add noise to $z_{t+1}$. This is because the model will collapse the latent space to inflate $\ell_\cons$ as shown in Table~\ref{table:ablation}, leading to a degenerate solution. Adding noise to $z_{t+1}$ prevents the map from collapsing; since the noise variance is fixed, $\ell_\cpc$ is only maximized by pushing points apart. Indeed, \Cref{table:ablation} shows that when noise is added but $\ell_\cons$ is removed, the latent map expands aggressively.

\textit{Curvature:} Finally, as previously observed in~\cite{levine2020prediction}, imposing low curvature is an important component if we want to use locally-linear control algorithms such as iLQR. Without the curvature loss, the Taylor approximation when running iLQR might not be accurate, leading to poor control performance. The right-most map in Figure~\ref{figure:map_ablation} shows a depiction of this setting, in which the map has very sharp regions, requiring the transition function to have high-curvature to move in these regions.

\subsection{Control Performance Comparison}

\textbf{Experimental pipeline:} For each control domain, we run $10$ different subtasks (different initial and/or goal states), and report the average performance among these subtasks. For PCC and PC3, we train $10$ different models, and each of them will perform all $10$ subtasks (which means a total of $10 \times 10 = 100$ subtasks), and we additionally report the performance of the best model. SOLAR training procedure depends on the specific subtask (i.e., initial and goal state), and since we cannot train $100$ different models due to huge computation cost, we train only 1 model for each subtask. All subtasks are shared for three methods.

\textbf{Result:} Table~\ref{table:control_compare} shows that our proposed PC3 model significantly outperforms the baselines by comparing the means and standard error of means on the different control tasks.\footnote{Due to huge computation cost of SOLAR, we lower control horizon for Balance, Swing Up, Cartpole and 3-Link, compared to what was used in the PCC paper.} PCC and SOLAR often fail at difficult tasks such as Swing Up and 3-Link. Moreover, SOLAR training procedure depends on the specific task, which makes them unsuitable to be reused for different tasks in the same environment. Figure~\ref{figure:map_compare} demonstrates some (randomly selected) latent maps of Planar and Inverted Pendulum domains learned by PCC and PC3. In general, PC3 produces more interpretable latent representation for Pendulum, due to the fact that next observation prediction is too conservative and may force a model to care about things that do not matter to downstream tasks. Finally, in terms of computation, PC3 enjoys huge improvements over the baselines, with $1.85\times$ faster than PCC and $52.8\times$ faster than SOLAR.


\section{Related Work}

\textbf{LCE Approaches.} In contrast to existing LCE methods \cite{watter2015embed,banijamali2017robust,levine2020prediction,zhang2019solar,hafner2018learning}, our main contribution in PC3 is the development of an information-theoretic approach for minimizing the predictive suboptimality of the encoder and circumventing the need to perform explicit next-observation prediction. In particular, PC3 can be seen as a natural information-theoretic extension of PCC \cite{levine2020prediction}, which itself extended and improved upon E2C \cite{watter2015embed} and RCE \cite{banijamali2017robust}. Compared to SOLAR \cite{zhang2019solar}, PC3 (as well as PCC, RCE, and E2C) decouples the representation learning and latent dynamics estimation from control---once the encoder and latent dynamics have been learned for a particular environment, it can be used to solve many SOC problems within the same environment. In contrast, SOLAR is an online algorithm that interleaves model learning with policy optimization. Furthermore, the latent model in SOLAR is restricted to be globally linear, which can potentially impact the control performance.

\textbf{Information-Theoretic Approaches.} Several works have previously explored information-theoretic approaches for representation learning in the reinforcement learning context \cite{nachum2018near,anand2019unsupervised,lu2019predictive}. However, these works do not test the quality of their learned representations for the purposes of model-based planning in the latent space, opting instead to leverage the representations for model-free RL. This is particularly notable in the case of \citet{nachum2018near}, who explicitly learned both an encoder $E$ and latent dynamics model $F$. As we showed in \cref{subsec:enforce_consistency}, maximizing the CPC bound alone may not be sufficient for ensuring that $F$ is a good predictor of the latent dynamics induced by $(p, E)$. Thus, the resulting $(E, F)$ from predictive coding alone may be unsuitable for multi-step latent planning, as we demonstrate in our ablation analysis in \cref{subsec:ablation}. 

\section{Conclusion}

In this work, we propose a novel information-theoretic Learning Controllable Embedding approach for handling high-dimensional stochastic optimal control. Our approach challenges the necessity of the next-observation prediction in existing LCE algorithms. We show theoretically that predictive coding is a valid alternative to next-observation prediction for learning a representation that minimizes predictive suboptimality. To instantiate information-theoretic LCE, we develop the Predictive Coding-Consistency-Curvature (PC3) model and show that PC3 is a simpler, yet more effective method than existing next-observation prediction-based LCE approaches. 
We also provide a thorough study on various components of the PC3 objective via ablation analysis to assist the adoption of predictive coding in future LCE research. 
A natural follow-up would be to study the efficacy of predictive coding when used in conjunction with other techniques in the LCE literature (e.g. latent overshooting) as well as with other controllers beyond the class of locally-linear controllers considered in our present work.


\bibliography{main}
\bibliographystyle{icml2020}
\input{1000_appendix.tex}

\appendix
\onecolumn

\end{document}

%% file: figures/model.tex
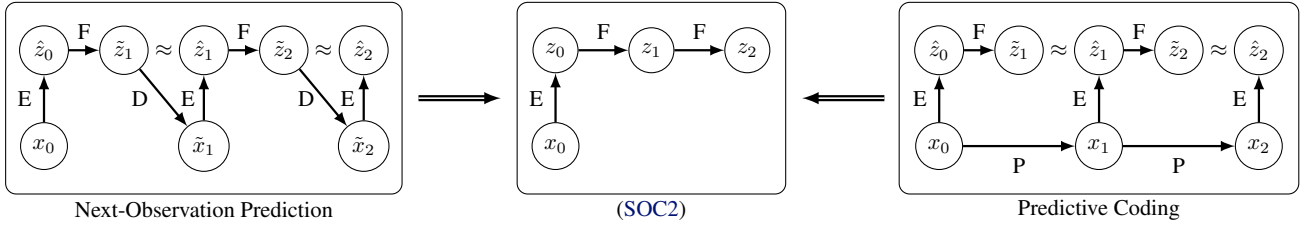
\begin{figure*}[ht]
\centering
\begin{tikzpicture}[scale=0.8, every node/.style={scale=0.85}]
\node[circ] (lz0) {$\hat{z}_0$};
\node[circ, xshift=1.25cm, yshift=0cm] (lz1t) {$\tilde{z}_1$};
\node[circ, xshift=2.5cm, yshift=0cm] (lz1) {$\hat{z}_1$};
\node[circ, xshift=3.75cm, yshift=0cm] (lz2t) {$\tilde{z}_2$};
\node[circ, xshift=5cm, yshift=0cm] (lz2) {$\hat{z}_2$};
\node[circ, xshift=0cm, yshift=-1.5cm] (lx0) {$x_0$};
\node[circ, xshift=2.5cm, yshift=-1.5cm] (lx1) {$\tilde{x}_1$};
\node[circ, xshift=5cm, yshift=-1.5cm] (lx2) {$\tilde{x}_2$};
\node[none, xshift=-0.3cm, yshift=-0.75cm] {E};
\node[none, xshift=1.5cm, yshift=-0.75cm] {D};
\node[none, xshift=2.25cm, yshift=-0.75cm] {E};
\node[none, xshift=4.1cm, yshift=-0.75cm] {D};
\node[none, xshift=4.75cm, yshift=-0.75cm] {E};
\node[none, xshift=0.625cm, yshift=0.3cm] {F};
\node[none, xshift=3.125cm, yshift=0.3cm] {F};
\node[none, xshift=1.875cm, yshift=0.0cm] {$\approx$};
\node[none, xshift=4.375cm, yshift=0.0cm] {$\approx$};
\node[draw, rectangle, rounded corners, minimum width=6.2cm, minimum height=3cm, xshift=2.5cm, yshift=-0.75cm] {};
\node[none, xshift=2.5cm, yshift=-2.5cm] () {Next-Observation Prediction};
\node[none, xshift=16.5cm, yshift=-2.5cm] () {Predictive Coding};
\node[none, xshift=9.5cm, yshift=-2.5cm] () {\eqref{problem:soc2}};

\node[circ, xshift=8cm, yshift=0cm] (mz0) {$z_0$};
\node[circ, xshift=9.5cm, yshift=0cm] (mz1) {$z_1$};
\node[circ, xshift=11cm, yshift=0cm] (mz2) {$z_2$};
\node[circ, xshift=8cm, yshift=-1.5cm] (mx0) {$x_0$};
\node[none, xshift=7.7cm, yshift=-0.75cm] {E};
\node[none, xshift=8.75cm, yshift=0.3cm] {F};
\node[none, xshift=10.25cm, yshift=0.3cm] {F};
\node[draw, rectangle, rounded corners, minimum width=4.2cm, minimum height=3cm, xshift=9.5cm, yshift=-0.75cm] {};

\node[circ, xshift=14cm, yshift=0cm] (rz0) {$\hat{z}_0$};
\node[circ, xshift=15.25cm, yshift=0cm] (rz1t) {$\tilde{z}_1$};
\node[circ, xshift=16.5cm, yshift=0cm] (rz1) {$\hat{z}_1$};
\node[circ, xshift=17.75cm, yshift=0cm] (rz2t) {$\tilde{z}_2$};
\node[circ, xshift=19cm, yshift=0cm] (rz2) {$\hat{z}_2$};
\node[circ, xshift=14cm, yshift=-1.5cm] (rx0) {$x_0$};
\node[circ, xshift=16.5cm, yshift=-1.5cm] (rx1) {$x_1$};
\node[circ, xshift=19cm, yshift=-1.5cm] (rx2) {$x_2$};
\node[none, xshift=13.7cm, yshift=-0.75cm] {E};
\node[none, xshift=16.2cm, yshift=-0.75cm] {E};
\node[none, xshift=18.7cm, yshift=-0.75cm] {E};
\node[none, xshift=14.625cm, yshift=0.3cm] {F};
\node[none, xshift=17.125cm, yshift=0.3cm] {F};
\node[none, xshift=15.875cm, yshift=0.0cm] {$\approx$};
\node[none, xshift=18.375cm, yshift=0.0cm] {$\approx$};
\node[none, xshift=15.25cm, yshift=-1.8cm] {P};
\node[none, xshift=17.75cm, yshift=-1.8cm] {P};
\node[draw, rectangle, rounded corners, minimum width=6.25cm, minimum height=3cm, xshift=16.5cm, yshift=-0.75cm] {};

\node[circ, xshift=5.5cm, yshift=-0.75cm, draw=none] (l) {};
\node[circ, xshift=7.5cm, yshift=-0.75cm, draw=none] (ml) {};
\node[circ, xshift=11.5cm, yshift=-0.75cm, draw=none] (mr) {};
\node[circ, xshift=13.5cm, yshift=-0.75cm, draw=none] (r) {};

\path
(lz0) edge [connect] (lz1t)
(lz1) edge [connect] (lz2t)
(lx0) edge [connect] (lz0)
(lz1t) edge [connect] (lx1)
(lz2t) edge [connect] (lx2)
(lx1) edge [connect] (lz1)
(lx2) edge [connect] (lz2)

(mz0) edge [connect] (mz1)
(mz1) edge [connect] (mz2)
(mx0) edge [connect] (mz0)

(rx0) edge [connect] (rx1)
(rx1) edge [connect] (rx2)
(rx0) edge [connect] (rz0)
(rx1) edge [connect] (rz1)
(rx2) edge [connect] (rz2)
(rz0) edge [connect] (rz1t)
(rz1) edge [connect] (rz2t)

(l) edge [doubleconnect] (ml)
(r) edge [doubleconnect] (mr)
;
\end{tikzpicture}  
\caption{Two high-level approaches to learn an $E$ and $F$ to instantiate \eqref{problem:soc2}. One way is to explicitly introduce a decoder $D$ and do next-observation prediction (left), whereas our method uses $F$ as a variational device to train $E$ via predictive coding (right).}
\label{fig:model}
\end{figure*}

%% file: figures/map_ablation.tex
\begin{figure*}[!ht]
    \vskip 0.1in
    \centering
    \begin{subfigure}[h]{0.2\textwidth}
        \centering
        \includegraphics[width=\textwidth]{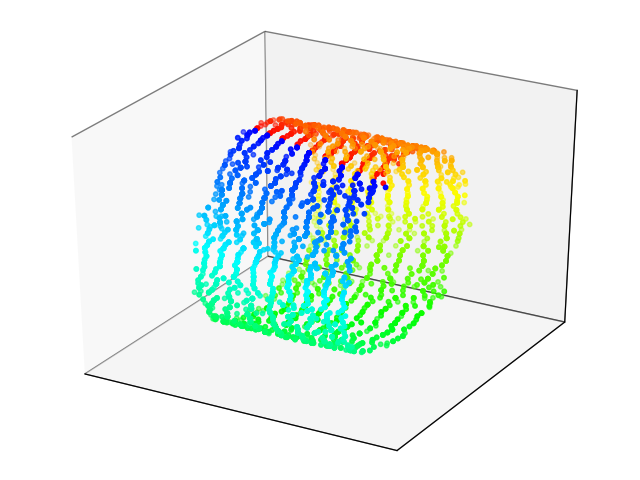}
    \end{subfigure}%
    \begin{subfigure}[h]{0.2\textwidth}
        \centering
        \includegraphics[width=\textwidth]{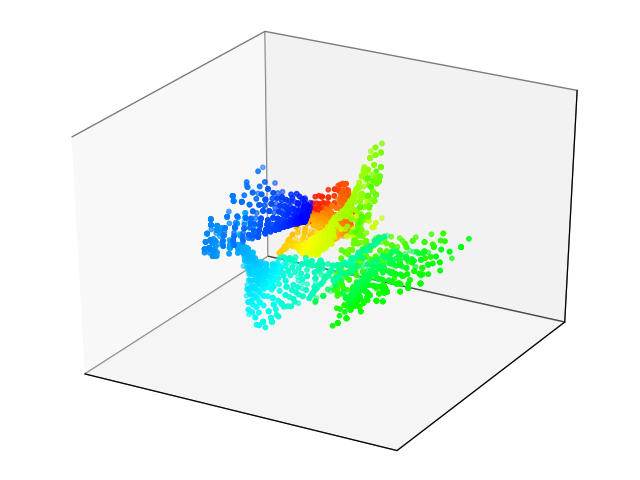}
    \end{subfigure}%
    \begin{subfigure}[h]{0.2\textwidth}
        \centering
        \includegraphics[width=\textwidth]{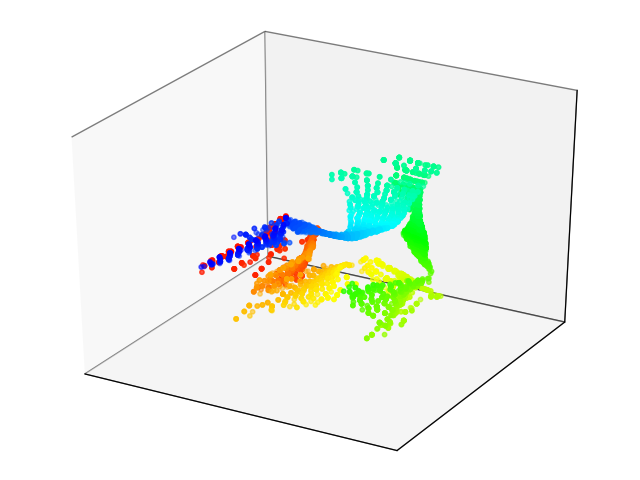}
    \end{subfigure}%
    \begin{subfigure}[h]{0.2\textwidth}
        \centering
        \includegraphics[width=\textwidth]{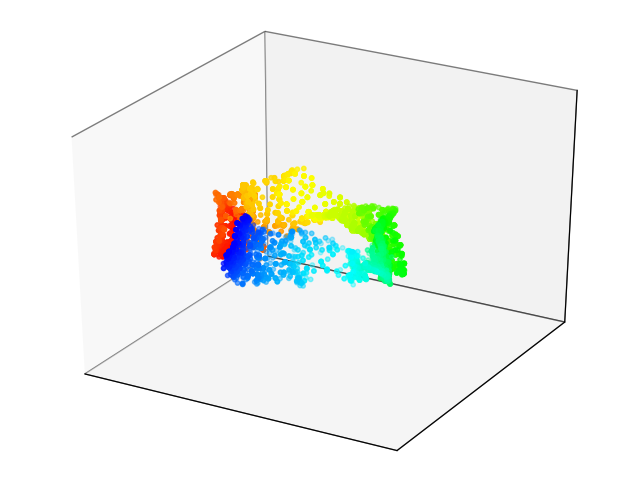}
    \end{subfigure}%
    \begin{subfigure}[h]{0.2\textwidth}
        \centering
        \includegraphics[width=\textwidth]{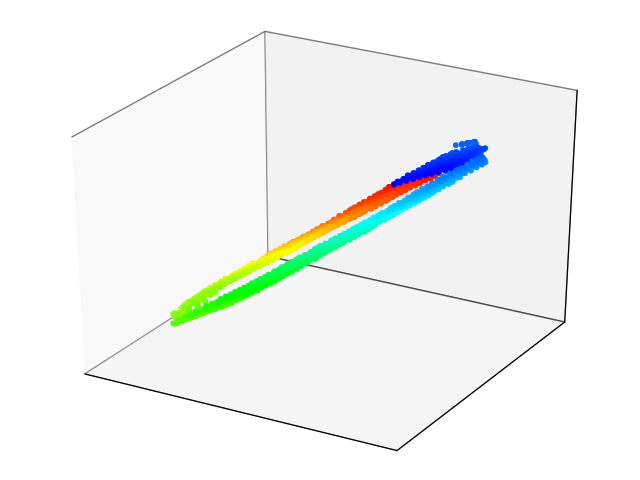}
    \end{subfigure}%
    \caption{Inverted pendulum representations. From left to right: PC3, w/o $(\ell_{\cons}, \eps)$, w/o $(\ell_\cons)$, w/o $\eps$, w/o $\ell_\textrm{curv}$}
    \label{figure:map_ablation}
    \vskip -0.1in
\end{figure*}

%% file: tables/ablation.tex
\begin{table*}[!ht]
    \centering
    \caption{Ablation analysis. Percentage of steps spent in goal state. From top to bottom: full-model PC3, excluding consistency and latent noise, excluding consistency, excluding latent noise, excluding curvature. For each setting we report the latent map scale, CPC, consistency and curvature loss, and control performance on balance and swing.}
    \label{table:ablation}
    \vskip 0.1in
    \begin{tabular}{ccccccc}
         \toprule
         Setting & Latent map size & $\ell_\cpc$ & $\ell_\cons$ & $\ell_\curv$ & Balance & Swing Up \\
         \midrule
         PC3 & $16.2$ & $4.58$ & $2.13$ & $0.03$ & $\mathbf{99.12 \pm 0.66}$ & $\mathbf{58.4 \pm 3.53}$\\
         w/o $(\ell_{\cons}, \epsilon)$ & $10.47$ & $5.07$ & $-4.13$ & $0.001$ & $34.55 \pm 3.69$ & $17.83 \pm 2.9$ \\
         w/o $\ell_\cons$ & $101.52$ & $5.03$ & $-4.87$ & $0.0025$ & $31.08 \pm 3.57$ & $7.46 \pm 1.32$\\
         w/o $\eps$ & $0.04$ & $3.27$ & $20.83$ & $0.0009$ & $65.2 \pm 1.11$ & $0 \pm 0$\\
         w/o $\ell_\textrm{curv}$ & $66.1$ & $4.8$ & $2.34$ & $0.56$ & $96.89 \pm 0.97$ & $21.69 \pm 2.73$\\
         \bottomrule
    \end{tabular}
    \vskip -0.1in
\end{table*}

%% file: tables/latent_prediction.tex
\begin{table*}[!ht]
    \centering
    \caption{We took a pretrained PC3 model, froze the encoder $E$, and then retrained only the latent dynamics model $F$ either without $\ell_\cons$ (first row) or without $\ell_\cpc$ (second row). Note that we continue to use $\ell_\curv$ and add $\eps$ noise in both settings.}
    \label{table:latent_prediction}
    \vskip 0.1in
    \begin{tabular}{ccccccc}
         \toprule
         Setting & Latent map size & $\ell_\cpc$ & $\ell_\cons$ & $\ell_\textrm{curv}$ & Balance & Swing Up \\
         \midrule
         Retrain $F$ w/o $\ell_\cons$ & $16.2$ & $4.57$ & $-21.93$ & $0.02$ & $46.77$ $\pm$ $3.66$ & $18.06 \pm 1.87$\\
         Retrain $F$ w/o $\ell_\cpc$ & $16.2$ & $4.6$ & $2.17$ & $0.03$ & $90.85$ $\pm$ $2.33$ & $50.11 \pm 3.74$ \\
         \bottomrule
    \end{tabular}
    \vskip -0.1in
\end{table*}

%% file: figures/map_compare.tex
\begin{figure*}[ht]
    \vskip 0.1in
    \centering
    \begin{subfigure}[h]{0.13\textwidth}
        \centering
        \includegraphics[width=\textwidth]{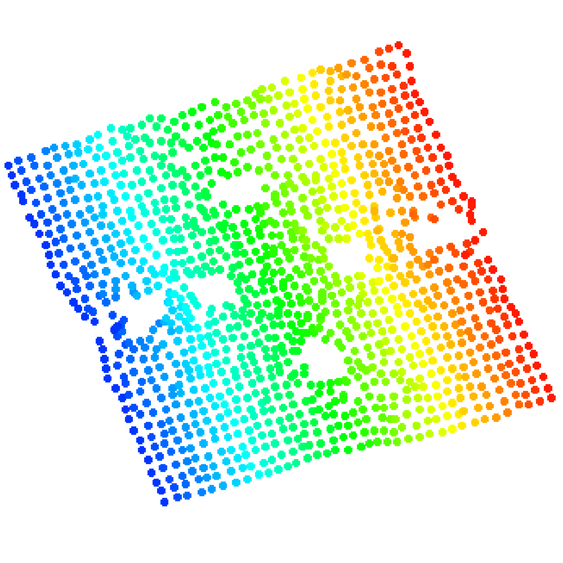}
    \end{subfigure}%
    ~~~~
    \begin{subfigure}[h]{0.13\textwidth}
        \centering
        \includegraphics[width=\textwidth]{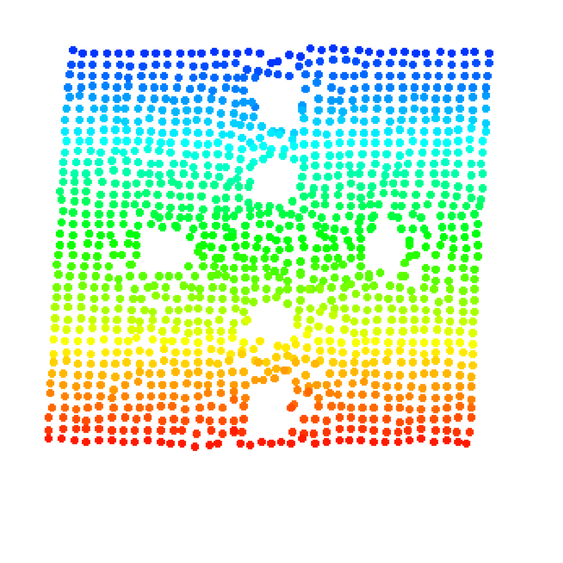}
    \end{subfigure}%
    ~~~~
    \begin{subfigure}[h]{0.13\textwidth}
        \centering
        \includegraphics[width=\textwidth]{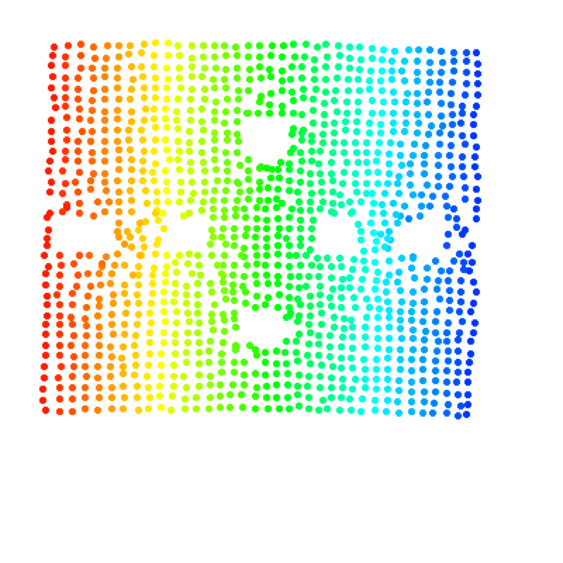}
    \end{subfigure}%
    ~~~~~~
    ~~~~~~
    \begin{subfigure}[h]{0.13\textwidth}
        \centering
        \includegraphics[width=\textwidth]{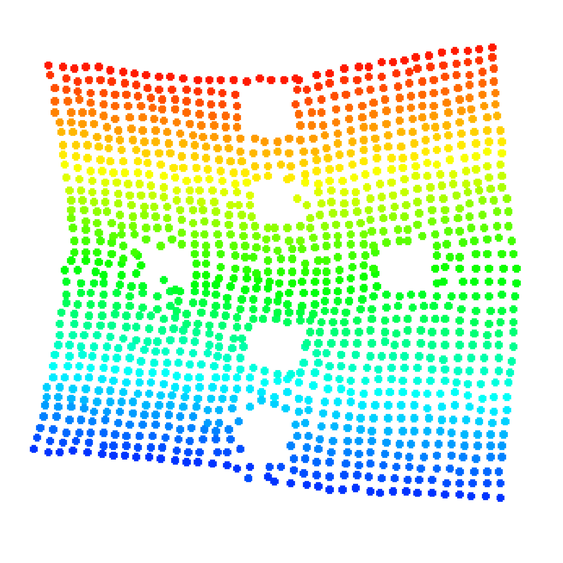}
    \end{subfigure}%
    ~~~~
    \begin{subfigure}[h]{0.13\textwidth}
        \centering
        \includegraphics[width=\textwidth]{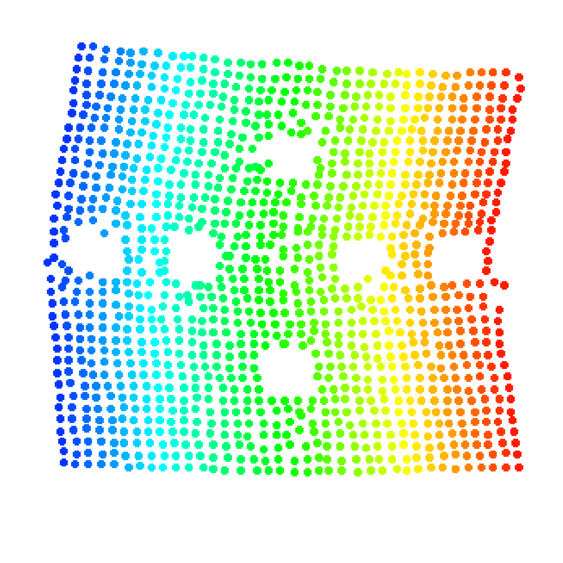}
    \end{subfigure}%
    ~~~~
    \begin{subfigure}[h]{0.13\textwidth}
        \centering
        \includegraphics[width=\textwidth]{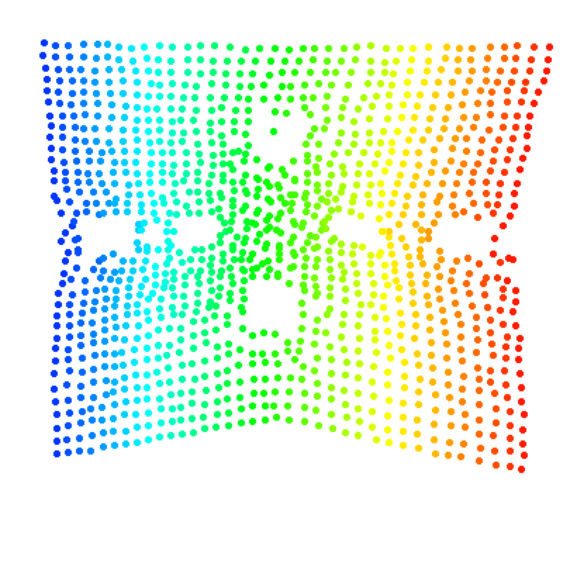}
    \end{subfigure}%
    
    \begin{subfigure}[h]{0.15\textwidth}
        \centering
        \includegraphics[width=\textwidth]{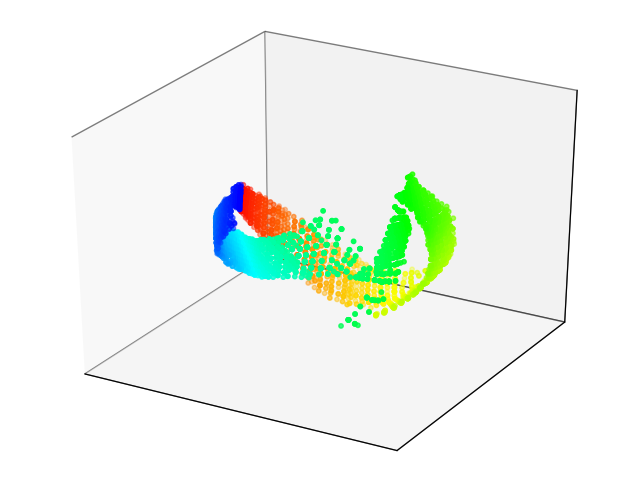}
        \caption*{}
    \end{subfigure}%
    \begin{subfigure}[h]{0.15\textwidth}
        \centering
        \includegraphics[width=\textwidth]{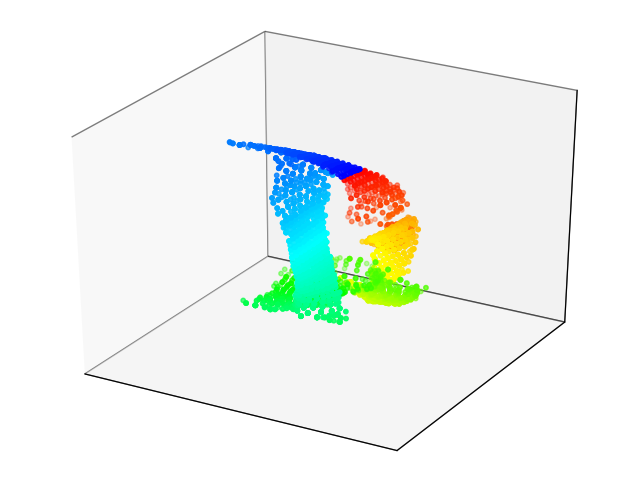}
        \caption*{PCC}
    \end{subfigure}%
    \begin{subfigure}[h]{0.15\textwidth}
        \centering
        \includegraphics[width=\textwidth]{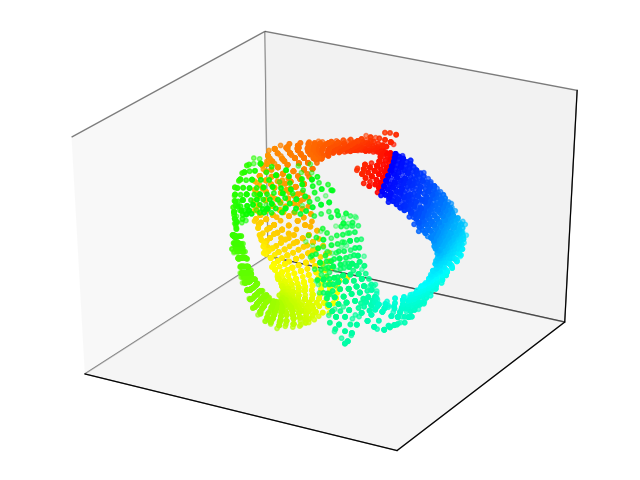}
        \caption*{}
    \end{subfigure}%
    ~~~~~
    ~~~~~
    \begin{subfigure}[h]{0.15\textwidth}
        \centering
        \includegraphics[width=\textwidth]{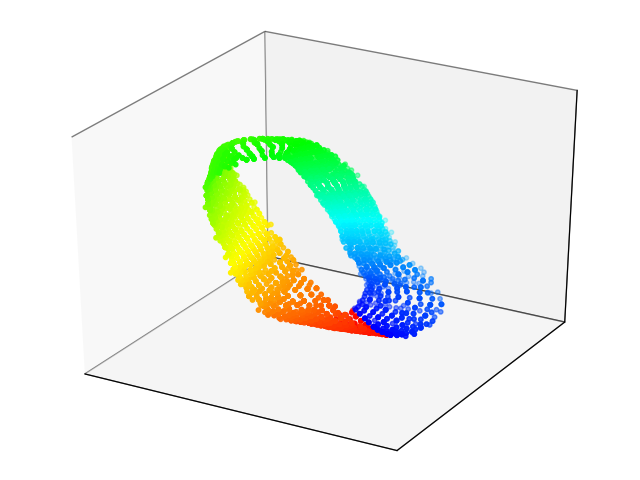}
        \caption*{}
    \end{subfigure}%
    \begin{subfigure}[h]{0.15\textwidth}
        \centering
        \includegraphics[width=\textwidth]{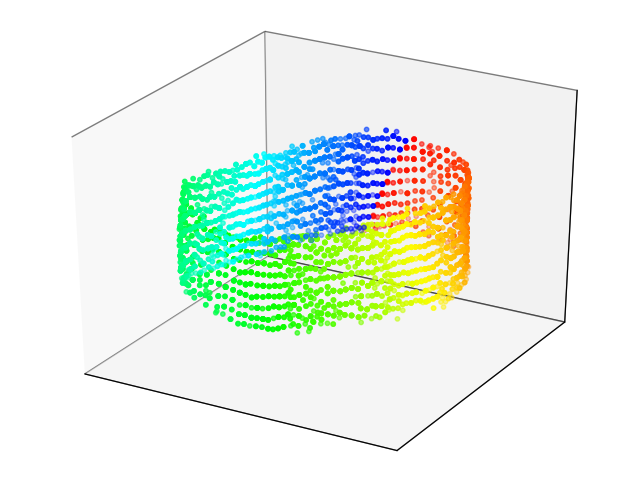}
        \caption*{PC3}
    \end{subfigure}%
    \begin{subfigure}[h]{0.15\textwidth}
        \centering
        \includegraphics[width=\textwidth]{latent_maps/pc3_7.png}
        \caption*{}
    \end{subfigure}%
    \centering
    \caption{Top: Planar latent representations; Bottom: Inverted Pendulum latent representations. Left three: PCC, right three: PC3.}
    \label{figure:map_compare}
    \vskip -0.1in
\end{figure*}

%% file: tables/control_compare.tex
\begin{table*}[ht]
    \centering
    \caption{Percentage steps in goal state for the average model (all) and top $1$ model. Since SOLAR is task-specific, it does not have top $1$.}
    \label{table:control_compare}
    \vskip 0.1in
    \begin{tabular}{cccccc}
         \toprule
         Task & PC3 (all) & PCC (all) & SOLAR (all) & PC3 (top $1$) & PCC (top $1$) \\
         \midrule
         Planar & $\mathbf{74.35 \pm 0.76}$ & $56.6 \pm 3.15$ & $68 \pm 3.8$ & $\mathbf{75.5 \pm 0.32}$ & $\mathbf{75.5 \pm 0.32}$ \\
         Balance & $\mathbf{99.12 \pm 0.66}$ & $91.9 \pm 1.72$ & $67 \pm 2.6$ & $\mathbf{100 \pm 0}$ & $\mathbf{100 \pm 0}$ \\
         Swing Up & $\mathbf{58.4 \pm 3.53}$ & $26.41 \pm 2.64$ & $35.4 \pm 1.9$ & $\mathbf{84 \pm 0}$ & $66.9 \pm 3.8$ \\
         Cartpole & $\mathbf{96.26 \pm 0.95}$ & $94.44 \pm 1.34$ & $91.2 \pm 5.4$ & $\mathbf{97.8 \pm 1.4}$ & $\mathbf{97.8 \pm 1.4}$ \\
         3-link & $\mathbf{42.4 \pm 3.23}$ & $14.17 \pm 2.2$ & $0 \pm 0$ & $\mathbf{78 \pm 1.04}$ & $45.8 \pm 6.4$ \\
         \bottomrule
    \end{tabular}
    \vskip -0.1in
\end{table*}

%% file: 1000_appendix.tex
\clearpage
\appendix
\onecolumn
\icmltitle{Supplementary Materials to \sharedtitle}

\section{Proofs in Section \ref{sec:itlce}}

\subsection{Connecting \eqref{problem:soc1} and \eqref{problem:soc1-e} with Next-observation Prediction}\label{proof:soc1soc1-e}
Recall that for an arbitrarily given encoder $E$ the proxy cost function in the observation space is given by $c_E(x, u) := \Expect\Big[\bar c(z, u) \mid \enc(x)\Big]$, where $z$ is sampled from $E(x)$. Equipped with this cost the only  difference between \eqref{problem:soc1-e}, i.e., 
$\min_{\Control} \; L(\Control, p, c_E,\obs_0)$,
and the original problem \eqref{problem:soc1}, i.e.,
$\min_{\Control} \; L(\Control, p, c,\obs_0)$,
is on the cost function used.

To motivate the heuristic method of learning an encoder $E$ by maximizing the likelihood of the next-observation prediction model, we want to show there exists at least one latent cost function $\bar c$ such that the aforementioned approach makes sense.
Followed from the equivalence of the energy-based graphical model (Markov random field) and Bayesian neural network \cite{koller2009probabilistic}, for any arbitrary encoder $E$ there exists a latent dynamics model $\tilde F$ and decoder $\tilde\dec$ such that any energy-based LCE model that has an encoder model $E$, namely $q_E(x'|x,u)$, can be written as $ (\tilde\dec\circ\tilde\dyn\circ\enc)(x'|x,u)$. 

Now, suppose for simplicity the observation cost is only state-dependent, and the latent cost $\bar c$ is constructed as follows: $\bar c(z,u):=\int_{x'}\int_{z'}c(x')d\tilde\dyn(z'|z,u)d\tilde\dec(x'|z')$. Then one can write $c_E(x,u)=\int_{x'}dq_E(x'|x,u) c(x')$, and this implies
\[
\left|\mathbb E_{x'\sim p(\cdot|x,u)}[c(x')]-c_E(x,u)\right|\leq c_{\max}\cdot D_{\text{TV}}(p(\cdot|x,u)||q_E(\cdot|x,u)),
\]
where $D_{\text{TV}}$ is the total variation distance of two distributions. 
Using analogous derivations of Lemma 11 in \cite{ghavamzadeh2016safe}, for the case of finite-horizon MDPs, one has the following chain of inequalities for any given control sequence $\{u_t\}_{t=0}^{T-1}$ and initial observation $\obs_0$:
\[
\begin{split}
|L(\Control, p,c,\obs_0)-L(\Control,p,c_E,\obs_0)|=&\left|\mathbb E\left[\sum_{t=1}^{T}c_{t}(\obs_t)\!\mid\! P,\obs_0\right]-\mathbb E\left[\sum_{t=0}^{T-1}c_{E,t}(\obs_t,\control_t)\!\mid\! P,\obs_0\right]\right|\\
\leq& T^2 \cdot c_{\max}\, \mathbb E\left[\frac{1}{T}\sum_{t=0}^{T-1}D_{\text{TV}}(p(\cdot|\obs_t,\control_t)||q_E(\cdot|\obs_t,\control_t))\mid P,\obs_0\right]\\
\leq&\sqrt{2}T^2\cdot c_{\max}\, \mathbb E\left[\frac{1}{T}\sum_{t=0}^{T-1}\sqrt{D_\text{KL}(p(\cdot|\obs_t,\control_t)|| \widehat{p}_E(\cdot|\obs_t,\control_t))}\mid P,\obs_0\right]\\
\leq&\sqrt{2} T^2\cdot c_{\max}\, \sqrt{\mathbb E\left[\frac{1}{T}\sum_{t=0}^{T-1}D_\text{KL}(p(\cdot|\obs_t,\control_t)||\widehat{p}_E(\cdot|\obs_t,\control_t))\mid P,\obs_0\right]},
\end{split}
\]
The first inequality is based on the result of the above lemma, the second inequality is based on Pinsker's inequality, and the third inequality is based on Jensen's inequality of $\sqrt{(
\cdot)}$ function.

Notice that for any arbitrary action sequence it can always be expressed in form of deterministic policy $u_t=\pi'(x_t,t)$ with some non-stationary state-action mapping $\pi'$. Therefore, the KL term can be written as:
\begin{equation}\label{eq:union_bound}
\begin{split}
&\mathbb E\left[\frac{1}{T}\sum_{t=0}^{T-1}D_\text{KL}(p(\cdot|\obs_t,\control_t)||q_E(\cdot|\obs_t,\control_t))\mid p,\pi,\obs_0\right]\\
=& \mathbb E\left[\frac{1}{T}\sum_{t=0}^{T-1}\int D_\text{KL}(p(\cdot|\obs_t,\control_t)||q_E(\cdot|\obs_t,\control_t))d\pi'(\control_t|\obs_t,t)\mid p,\obs_0\right] \\
=&\mathbb E\left[\frac{1}{T}\sum_{t=0}^{T-1}\int D_\text{KL}(p(\cdot|\obs_t,\control_t)||q_E(\cdot|\obs_t,\control_t))\cdot \frac{d\pi'(\control_t|\obs_t,t)}{d\Control(\control_t)}\cdot d\Control(\control_t)\mid p,\obs_0\right] \leq  \overline\Control\cdot\mathbb E_{\obs,\control}\left[D_\text{KL}(p(\cdot|\obs,\control)||q_E(\cdot|\obs,\control))\right],
\end{split}
\end{equation}
where the expectation is taken over the state-action stationary distribution of the finite-horizon problem that is induced by data-sampling policy $\Control$.
The last inequality is due to change of measures in policy, and the last inequality is due to the facts that (i) $\pi$ is a deterministic policy, (ii) $d\Control(\control_t)$ is a sampling policy with lebesgue measure $1/\overline{\Control}$ over all control actions, (iii) the following bounds for importance sampling factor holds: $\left|\frac{d\pi'(\control_t|\obs_t,t)}{d\Control(\control_t)}\right|\leq\overline{\Control}$. 

Combining the above arguments we have the following inequality for any given encoder model $E$ and any control sequence $U$:
\begin{equation}\label{eq:ineq_1}
|L(\Control, p,c,\obs_0)-L(\Control,p,c_E,\obs_0)|\leq  \sqrt{2} T^2\cdot c_{\max}\overline\Control\cdot\sqrt{\mathbb E_{\obs,\control}\left[D_\text{KL}(p(\cdot|\obs,\control)||q_E(\cdot|\obs,\control))\right]}.
\end{equation}
Using the above results we now have the following sub-optimality performance bound between the optimizer of \eqref{problem:soc1}, $\Control^*_1$, and the optimizer of \eqref{problem:soc1-e}, $\Control^*_{\text{1-E}}$:
\begin{equation}
\begin{split}
L(\Control^*_1, p,c,\obs_0)\ge & L(\Control^*_1,p,c_E,\obs_0)-\sqrt{2} T^2\cdot c_{\max}\overline\Control\cdot\sqrt{\mathbb E_{\obs,\control}\left[D_\text{KL}(p(\cdot|\obs,\control)||q_E(\cdot|\obs,\control))\right]}\\
\geq &L(\Control^*_{\text{1-E}},p,c_E,\obs_0)-\sqrt{2} T^2\cdot c_{\max}\overline\Control\cdot\sqrt{\mathbb E_{\obs,\control}\left[D_\text{KL}(p(\cdot|\obs,\control)||q_E(\cdot|\obs,\control))\right]}.
\end{split}
\end{equation}

This shows that the performance gap between \eqref{problem:soc1} and \eqref{problem:soc1-e} is bounded by the prediction loss  $\sqrt{2} T^2\cdot c_{\max}\overline\Control\cdot \sqrt{\mathbb E_{\obs,\control}\left[D_\text{KL}(p(\cdot|\obs,\control)||q_E(\cdot|\obs,\control))\right]}$. Thus this result motivates the approach of learning the encoder model $E$ of proxy cost by maximizing the likelihood of the next-observation prediction LCE model.

\newpage
\subsection{Proof of Lemma \ref{lem:pred_subopt}}\label{proof:pred_subopt}
We first provide the proof in a more general setting. Consider the data distribution $p(x, y)$. Given any two representation functions $e: \X \to \A$ and $f: \Y \to \B$, we wish to inquire how good these two functions are for constructing a predictor of $y$ given $x$. To do so, we introduce a restricted class of prediction models of the form
\begin{align}
    q_\psi(y \giv x) \propto \psi_1(y)\psi_2(e(x), f(y)),
\end{align}
Let $q^*(y \mid x)$ denote the model that minimizes 
\begin{align}
    \ell^* = \min_q \Expect_{p(x)} D_{KL}({p(y \mid x)}||{q_\psi(y \mid x)}).
\end{align}
Our goal is to upper bound the best possible loss $\ell^*$ based on the mutual information gap $I(X \scolon Y) - I( e(X) \scolon f(Y))$. In particular, we find that
\begin{align}
    \Expect_{p(x)} D_{KL}({p(y \mid x)}||{q^*(y \mid x)}) \le
    I(X \scolon Y) - I( e(X) \scolon f(Y)).
\end{align}
We prove via explicit construction of a model $q(y \giv x)$ whose corresponding loss $\ell$ is exactly the mutual information gap. Let $(X, Y)$ be joint random variables associated with $p(x, y)$. Let $r(a \giv b)$ be the conditional distribution of $a = e(x)$ given $b = f(y)$ associated with the joint random variables $(A, B) = (e(X), f(Y))$. Simply choose
\begin{align}
    q(y \giv x) \propto p(y)r(e(x)\giv f(y)) \implies
    q(y \giv x) 
    = \frac{p(y)r(e(x)\giv f(y))}{\Expect_{p(y')} r(e(x)\giv f(y'))}.
\end{align}
Then, by law of the unconscious statistician, we see that
\begin{align}
    \Expect_{p(x, y)}\ln q(y \giv x)
    &= -H(Y) + 
    \Expect_{p(x, y)} 
    \ln \frac{r(e(x)\giv f(y))}
    {\Expect_{p(y')} r(e(x)\giv f(y'))}\\
    &= -H(Y) + 
    \Expect_{r(a, b)}
    \ln \frac{r(a\giv b)}
    {\Expect_{r(b')} r(a\giv b')}\\
    &= -H(Y) + I_r(A \scolon B)\\
    &= -H(Y) + I(e(X) \scolon f(Y)).
\end{align}
Finally, we see that
\begin{align}
    \ell = \Expect_{p(x)} D_{KL}({p(y \giv x)}||{q(y \giv x)}) 
    &= -H(Y \giv X) - \Expect_{p(x, y)} \ln q(y \giv x)\\
    &= H(Y) - H(Y \giv X) - I(e(X) \scolon f(Y))\\
    &= I(X \scolon Y) - I(e(X) \scolon f(Y)).
\end{align}
Since $\ell^* \le \ell$, the mutual information gap thus upper bounds the loss associated with the best restricted predictor $q^*$.

To complete the proof for \cref{lem:pred_subopt}, simply let 
\begin{align}
    X &:= (X_t, U_t)\\
    Y &:= X_\tpo\\
    e(X) &:= (E(X_t), U_t)\\
    f(Y) &:= E(X_\tpo).
\end{align}

\newpage
\subsection{Proof of Lemma \ref{lem:consistency}}\label{proof:consistency}
For the first part of the proof, at any time-step $t\geq 1$, for any arbitrary control action sequence $\{u_t\}_{t=0}^{T-1}$, and any arbitrary latent dynamics model $F$, with a given encoder $\enc$ consider the following decomposition of the expected cost: 
$
\mathbb E[c(\obs_t,\control_t)\mid  P,\obs_0]=\mathbb E[\overline c(\latent_t,\control_t)\mid  \enc,P,\obs_0]=\int_{\obs_{0:t}}\prod_{k=1}^{t} P(\obs_k|\obs_{k-1},\control_{k-1})\cdot\int_{\latent_t}\enc(\latent_t|\obs_t)\bar c(\latent_t,\control_t).
$
Now consider the two-stage cost function: $\mathbb E[c(\obs_{t-1},\control_{t-1})+c(\obs_t,\control_t)\mid P,\obs_0]$. One can express this cost function as
\[
\begin{split}
&\mathbb E[\overline c(\latent_{t-1},\control_{t-1})+\overline c(\latent_t,\control_t)\mid  \enc,P,\obs_0]\\
=&\int_{\obs_{0:t-1}}\prod_{k=1}^{t-1} P(\obs_k|\obs_{k-1},\control_{k-1})\cdot \left(\int_{\latent_{t-1}}\enc(\latent_{t-1}|\obs_{t-1})\bar c(\latent_{t-1},\control_{t-1})+\int_{\obs_{t}}P(\obs_{t}|\obs_{t-1},\control_{t-1})\int_{\latent_{t}}\enc(\latent_{t}|\obs_{t})\bar c(\latent_{t},\control_{t})\right)\\
\leq &\int_{\obs_{0:t-2}}\prod_{k=1}^{t-2} P(\obs_k|\obs_{k-1},\control_{k-1}) \cdot\left(\int_{\latent_{t-2}}\enc(\latent_{t-2}|\obs_{t-2})\int_{\latent_{t-1}}\dyn(\latent_{t-1}|\latent_{t-2},\control_{t-2})\bar c(\latent_{t-1},\control_{t-1})\right.\\
&\left.+\int_{\obs_{t-1}}P(x_{t-1}|x_{t-2},u_{t-2})\int_{\latent_{t-1}}\enc(\latent_{t-1}|\obs_{t-1})\int_{\latent_t}\dyn(\latent_t|\latent_{t-1},\control_{t-1})\bar c(\latent_{t},\control_{t})\right) \\
&+c_{\max}\cdot \int_{\obs_{0:t-2}}\prod_{k=1}^{t-2} P(\obs_k|\obs_{k-1},\control_{k-1})\cdot \Big(D_{\text{TV}}\left(\enc\circ P(\cdot|x_{t-2},u_{t-2})||F\circ \enc(\cdot|x_{t-2},u_{t-2})\right)\\
&\qquad\qquad\qquad\qquad\qquad+\mathbb E_{x_{t-1}\sim P(\cdot|x_{t-2},u_{t-2})}\left[D_{\text{TV}}\left(\enc\circ P(\cdot|x_{t-1},u_{t-1})||F\circ \enc(\cdot|x_{t-1},u_{t-1})\right)\right]\Big)\\
\leq &\int_{\obs_{0:t-2}}\prod_{k=1}^{t-2} P(\obs_k|\obs_{k-1},\control_{k-1}) \int_{\latent_{t-2}}\enc(\latent_{t-2}|\obs_{t-2})\int_{\latent_{t-1}}\dyn(\latent_{t-1}|\latent_{t-2},\control_{t-2})\cdot\\
&\qquad\qquad\qquad\qquad\qquad\left(\bar c(\latent_{t-1},\control_{t-1})+\int_{\latent_t}\dyn(\latent_t|\latent_{t-1},\control_{t-1})\bar c(\latent_{t},\control_{t})\right)\\
&+c_{\max}\cdot \int_{\obs_{0:t-2}}\prod_{k=1}^{t-2} P(\obs_k|\obs_{k-1},\control_{k-1})\cdot \left(2\cdot D_{\text{TV}}\left(\enc\circ P(\cdot|x_{t-2},u_{t-2})||F\circ \enc(\cdot|x_{t-2},u_{t-2})\right)\right.\\
&\left.\qquad\qquad\qquad\qquad\qquad\qquad+\mathbb E_{x_{t-1}\sim P(\cdot|x_{t-2},u_{t-2})}\left[D_{\text{TV}}\left(\enc\circ P(\cdot|x_{t-1},u_{t-1})||F\circ \enc(\cdot|x_{t-1},u_{t-1})\right)\right]\right)\\
\end{split}
\]
The last inequality is based on the chain of inequalities at any $(x_{t-2}, u_{t-2})\in\mathcal X\times\mathcal U$:
\[
\begin{split}
&D_{\text{TV}}\left(\enc\circ P\circ P(\cdot|x_{t-2},u_{t-2})||F\circ F\circ \enc(\cdot|x_{t-2},u_{t-2})\right)\\
\leq&  D_{\text{TV}}\left(F\circ \enc\circ P(\cdot|x_{t-2},u_{t-2})||F\circ F\circ \enc(\cdot|x_{t-2},u_{t-2})\right)\\
&\qquad\qquad\qquad\qquad+ D_{\text{TV}}\left(\enc\circ P\circ P(\cdot|x_{t-2},u_{t-2})||F\circ \enc\circ P(\cdot|x_{t-2},u_{t-2})\right)\\
\leq &D_{\text{TV}}\left(\enc\circ P(\cdot|x_{t-2},u_{t-2})||F\circ \enc(\cdot|x_{t-2},u_{t-2})\right)\\
&\qquad\qquad\qquad\qquad+\mathbb E_{x_{t-1}\sim P(\cdot|x_{t-2},u_{t-2})}\left[D_{\text{TV}}\left(\enc\circ P(\cdot|x_{t-1},u_{t-1})||F\circ \enc(\cdot|x_{t-1},u_{t-1})\right)\right],
\end{split}
\]
in which the first one is based on triangle inequality and the second one is based on the non-expansive property of $D_{TV}$.
By continuing the above expansion,  one can show that
\begin{equation}\label{eq:kl_tv}
\begin{split}
&\left|\mathbb E\left[L(\Control,F,\overline c,z_0)\mid \enc,x_0\right]-L(\Control, P,c,\obs_0)\right|\\
=&\left|\mathbb E\left[L(\Control,F,\overline c,z_0)\mid \enc,x_0\right]-L(\Control, P, \bar c\circ \enc,\obs_0)\right|\\
\leq& T^2 \cdot c_{\max}\, \mathbb E\left[\frac{1}{T}\sum_{t=0}^{T-1}D_{\text{TV}}((\enc\circ P)(\cdot|\obs_t,\control_t)||(\dyn\circ\enc)(\cdot|\obs_t,\control_t))\mid P,\obs_0\right]\\
\leq& T^2 \cdot c_{\max}\, \mathbb E\left[\frac{1}{T}\sum_{t=0}^{T-1}\mathbb E_{\obs_{t+1}\sim P(\cdot|\obs_{t},\control_{t})}\left[D_{\text{TV}}(\enc(\cdot|\obs_{t+1})||(\dyn\circ\enc)(\cdot|\obs_t,\control_t))\right]\mid P,\obs_0\right]\\
\leq &\sqrt{2}\cdot\mathbb E_{\obs,\control,\obs'\sim P(\cdot|x,u)}\Big[\sqrt{D_{\text{KL}}\Big(\enc(\cdot|x')||\big(F\circ E\big)(\cdot|x,u)\Big)}\Big]\\
\leq &\sqrt{2\cdot\mathbb E_{\obs,\control,\obs'\sim P(\cdot|x,u)}\Big[D_{\text{KL}}\Big(\enc(\cdot|x')||\big(F\circ E\big)(\cdot|x,u)\Big)\Big]},
\end{split}
\end{equation}
where the second inequality  is based on convexity of $D_{TV}$, the third inequality is based on Pinsker's inequality and the last inequality is based on Jensen's inequality of $\sqrt{(
\cdot)}$ function.

For the second part of the proof, one can show the following chain of inequalities for solution of \eqref{problem:soc1-e} and \eqref{problem:soc2}:
\begin{equation*}
\begin{split}
&L(\Control^*_{\text{1-E}}, P,\overline c\circ\enc,\obs_0)\\
\ge & \mathbb E\left[L(\Control^*_{\text{1-E}},\dyn, \overline c,z_0)\mid \enc,x_0\right]-T^2\cdot c_{\max}\overline\Control\cdot\sqrt{2\cdot\mathbb E_{\obs,\control,\obs'\sim P(\cdot|\obs,\control)}\left[D_{\text{KL}}(\enc(\cdot|\obs_{t+1})||(\dyn\circ\enc)(\cdot|\obs_t,\control_t))\right]}\\
= & \mathbb E\left[L(\Control^*_{\text{1-E}},\dyn,\overline c,z_0)\mid \enc,x_0\right]+T^2\cdot c_{\max}\overline\Control\cdot\sqrt{2\cdot\mathbb E_{\obs,\control,\obs'\sim P(\cdot|\obs,\control)}\left[D_{\text{KL}}(\enc(\cdot|\obs_{t+1})||(\dyn\circ\enc)(\cdot|\obs_t,\control_t))\right]}\\
&-2 T^2\cdot c_{\max}\overline\Control\cdot\sqrt{2\cdot\mathbb E_{\obs,\control,\obs'\sim P(\cdot|\obs,\control)}\left[D_{\text{KL}}(\enc(\cdot|\obs_{t+1})||(\dyn\circ\enc)(\cdot|\obs_t,\control_t))\right]}\\
\geq & \mathbb E\left[L(\Control^*_{\text{2-EF}},\dyn,\overline c,z_0)\mid \enc,x_0\right]+ T^2\cdot c_{\max}\overline\Control\cdot\sqrt{2\cdot\mathbb E_{\obs,\control,\obs'\sim P(\cdot|\obs,\control)}\left[D_{\text{KL}}(\enc(\cdot|\obs_{t+1})||(\dyn\circ\enc)(\cdot|\obs_t,\control_t))\right]}\\
&-2T^2\cdot c_{\max}\overline\Control\cdot\sqrt{2\cdot\mathbb E_{\obs,\control,\obs'\sim P(\cdot|\obs,\control)}\left[D_{\text{KL}}(\enc(\cdot|\obs_{t+1})||(\dyn\circ\enc)(\cdot|\obs_t,\control_t))\right]}\\
\geq &L(\Control^*_{\text{2-EF}}, P,\overline c\circ \enc,\obs_0)-2T^2\cdot c_{\max}\overline\Control\cdot\sqrt{2\cdot\mathbb E_{\obs,\control,\obs'\sim P(\cdot|\obs,\control)}\left[D_{\text{KL}}(\enc(\cdot|\obs_{t+1})||(\dyn\circ\enc)(\cdot|\obs_t,\control_t))\right]}\\
\geq &L(\Control^*_{\text{2-EF}}, P,c,\obs_0)-2\underbrace{T^2\cdot c_{\max}\overline\Control}_{\lambda_{\text{CON}}}\cdot\underbrace{\sqrt{2\cdot\mathbb E_{\obs,\control,\obs'\sim P(\cdot|\obs,\control)}\left[D_{\text{KL}}(\enc(\cdot|\obs_{t+1})||(\dyn\circ\enc)(\cdot|\obs_t,\control_t))\right]}}_{R_{\text{CON}}(\enc,\dyn)},
\end{split}
\end{equation*}
where the first and third inequalities are based on the first part of this lemma, and the second inequality is based on the optimality condition of problem \eqref{problem:soc2}.
This completes the proof.

\newpage
\section{Experiment Details}

In the following sections we will provide the description $4$ control domains and implementation details used in the experiments.

\subsection{Description of the domains}
All control environments are the same as reported in~\cite{levine2020prediction}, except that we report both balance and swing up tasks for pendulum, where the author only reported swing up.

\subsection{Implementation details}

\subsubsection{Hyperparameters}
SOLAR training specifics: We use their default setting:
\begin{itemize}
    \item Batch size of $2$.
    \item ADAM~\cite{kingma2014adam} with $\beta_1 = 0.9, \beta_2 = 0.999$, and $\epsilon = 10^{-8}$. Learning rate $\alpha_{\text{model}} = 2 \cdot 10^{-5} \times \text{horizon}$ for learning $\mathcal{MNIW}$ prior and $\alpha = 10^{-3}$ for other parameters.
    \item $(\beta_{\text{start}}, \beta_{\text{end}}, \beta_{\text{rate}}) = (10^{-4}, 10.0, 5 \cdot 10^{-5})$
    \item Local inference and control:
    \begin{itemize}
        \item Data strength: $50$
        \item KL step: $2.0$
        \item Number of rollouts per iteration: $100$
        \item Number of iterations: $10$
    \end{itemize}
\end{itemize}

PCC training specifics: We use their reported setting:
\begin{itemize}
    \item Batch size of $128$\footnote{Training with batch size of 256 gives worse results.}.
    \item ADAM with $\alpha = 5 \cdot 10^{-4}$, $\beta_1 = 0.9, \beta_2 = 0.999$, and $\epsilon = 10^{-8}$.
    \item L2 regularization with a coefficient of $10^{-3}$.
    \item $(\lambda_p, \lambda_c, \lambda_{\text{cur}}) = (1,8,8)$, and $\delta = 0.01$ for the curvature loss. This setting is shared across all domains.
    \item Additional VAE~\cite{kingma2013auto} loss term $\ell_{\text{VAE}} = -\mathbb{E}_{q(z|x)}[\log p(x|z)] + D_{\text{KL}}(q(z|x)||p(z))$ with a very small coefficient of $0.01$, where $p(z) = \mathcal{N}(0,1)$.
    \item Additional deterministic reconstruction loss with coefficient $0.3$: given the current observation $x$, we take the means of the encoder output and the dynamics model output, and decode to get the reconstruction of the next observation.
\end{itemize}

PC3 training specifics:
\begin{itemize}
    \item Batch size of $256$.
    \item ADAM with $\alpha = 5 \cdot 10^{-4}$, $\beta_1 = 0.9, \beta_2 = 0.999$, and $\epsilon = 10^{-8}$.
    \item L2 regularization with a coefficient of $10^{-3}$.
    \item Latent noise $\epsilon = 0.1$ and $\lambda_1 = 1$ across all domains without any tuning.
    \item $\lambda_2$ was set to be $1$ across all domains, after it was tuned using grid search in range $\{0.5,0.75,1\}$ on Planar system.
    \item $\lambda_3$ was set to be $7$ across all domains, after it was tuned using grid search in range $\{1,3,7\}$ on Planar system.
    \item $\delta=0.01$ for the curvature loss.
    \item Additional loss $\ell_{\text{add}} = ||\frac{1}{N}\sum_{i=1}^N z_i||_2^2$ with a very small coefficient of $0.01$, which is used to center the latent space around the origin. We found this term to be important to stabilize the training process.
\end{itemize}

\subsubsection{Network architectures}
We next present the specific architecture choices for each domain. For fair comparison, the architectures were shared across all algorithms when possible, ReLU non-linearities were used between each two layers.

\textbf{Encoder:} composed of a backbone (either a MLP or a CNN, depending on the domain) and an additional fully-connected (FLC) layer that outputs either a vector (for PC3) or a Gaussian distribution (for PCC and SOLAR).

\textbf{Latent dynamics (PCC and PC3):} the path that leads from $\{z,u\}$ to $z'$, composed of a MLP backbone and an additional FLC layer that outputs either a vector (for PC3) or a Gaussian distribution (for PCC and SOLAR).

\textbf{Decoder (PCC and SOLAR):} composed of a backbone (either a MLP or a CNN, depending on the domain) and an additional FLC layer that outputs a Bernoulli distribution.

\textbf{Backward dynamics:}  the path that leads from $\{z',u,x\}$ to $z$. Each of the inputs goes through a FLC network $\{N_z, N_u, N_x\}$, respectively. The outputs are concatenated and passed through another FLC network $N_{\text{joint}}$, and finally an additional FLC network which outputs a Gaussian distribution.

\textbf{Planar system}
\begin{itemize}
    \item Input: $40 \times 40$ images. $5000$ training samples of the form $(x,u,x')$ for PCC and PC3, and $125$ rollouts for SOLAR.
    \item Actions space: $2$-dimensional
    \item Latent space: $2$-dimensional
    \item Encoder: $3$ Layers: $300$ units - $300$ units - $4$ units for PCC and SOLAR ($2$ for mean and $2$ for variance) or $2$ units for PC3
    \item Dynamics: $3$ Layers: $20$ units - $20$ units - $4$ units for PCC and SOLAR or $2$ units for PC3
    \item Decoder: $3$ Layers: $300$ units - $300$ units - $1600$ units (logits)
    \item Backward dynamics: $N_z = 5, N_u = 5, N_x = 100 - N_{\text{joint}} = 100 - 4$ units
    \item Planning horizon: $T = 40$
    \item Initial standard deviation for collecting data (SOLAR): $1.5$ for both global and local traning.  
\end{itemize}

\textbf{Inverted Pendulum $-$ Swing up and Balance}
\begin{itemize}
    \item Input: Two $48 \times 48$ images. $20000$ training samples of the form $(x,u,x')$ for PCC and PC3, and $200$ rollouts for SOLAR.
    \item Actions space: $1$-dimensional
    \item Latent space: $3$-dimensional
    \item Encoder: $3$ Layers: $500$ units - $500$ units - $6$ units for PCC and SOLAR or $3$ units for PC3
    \item Dynamics: $3$ Layers: $30$ units - $30$ units - $4$ units for PCC and SOLAR or $2$ units for PC3
    \item Decoder: $3$ Layers: $500$ units - $500$ units - $4608$ units (logits)
    \item Backward dynamics: $N_z = 10, N_u = 10, N_x = 200 - N_{\text{joint}} = 200 - 6$ units
    \item Planning horizon: $T = 100$
    \item Initial standard deviation for collecting data (SOLAR): $0.5$ for both global and local training.  
\end{itemize}

\textbf{Cartpole}
\begin{itemize}
    \item Input: Two $80 \times 80$ images. $15000$ training samples of the form $(x,u,x')$ for PCC and PC3, and $300$ rollouts for SOLAR.
    \item Actions space: $1$-dimensional
    \item Latent space: $8$-dimensional
    \item Encoder: $6$ Layers: Convolutional layer: $32 \times 5 \times 5$; stride ($1$, $1$) - Convolutional layer: $32 \times 5 \times 5$; stride ($2$, $2$) - Convolutional layer: $32 \times 5 \times 5$; stride ($2$, $2$) - Convolutional layer: $10 \times 5 \times 5$; stride ($2$, $2$) - $200$ units - $16$ units for PCC and SOLAR or $8$ units for PC3
    \item Dynamics: $3$ Layers: $40$ units - $40$ units - $16$ units for PCC and SOLAR or $8$ units for PC3
    \item Decoder: $6$ Layers: $200$ units - $1000$ units - $100$ units - Convolutional layer: $32 \times 5 \times 5$; stride ($1$, $1$) - Upsampling ($2$, $2$) - Convolutional layer: $32 \times 5 \times 5$; stride ($1$, $1$) - Upsampling ($2$, $2$) - Convolutional layer: $32 \times 5 \times 5$; stride ($1$, $1$) - Upsampling ($2$, $2$) - Convolutional layer: $2 \times 5 \times 5$; stride ($1$, $1$)
    \item Backward dynamics: $N_z = 10, N_u = 10, N_x = 300 - N_{\text{joint}} = 300 - 16$ units
    \item Planning horizon: $T = 50$
    \item Initial standard deviation for collecting data (SOLAR): $10$ for global and $5$ for local training.
\end{itemize}

\textbf{3-link Manipulator $-$ Swing up}
\begin{itemize}
    \item Input: Two $80 \times 80$ images. $30000$ training samples of the form $(x,u,x')$ for PCC and PC3, and $150$ rollouts for SOLAR.
    \item Actions space: $3$-dimensional
    \item Latent space: $8$-dimensional
    \item Encoder: $6$ Layers: Convolutional layer: $32 \times 5 \times 5$; stride ($1$, $1$) - Convolutional layer: $32 \times 5 \times 5$; stride ($2$, $2$) - Convolutional layer: $32 \times 5 \times 5$; stride ($2$, $2$) - Convolutional layer: $10 \times 5 \times 5$; stride ($2$, $2$) - $200$ units - $16$ units for PCC and SOLAR or $8$ units for PC3
    \item Dynamics: $3$ Layers: $40$ units - $40$ units - $16$ units for PCC and SOLAR or $8$ units for PC3
    \item Decoder: $6$ Layers: $200$ units - $1000$ units - $100$ units - Convolutional layer: $32 \times 5 \times 5$; stride ($1$, $1$) - Upsampling ($2$, $2$) - Convolutional layer: $32 \times 5 \times 5$; stride ($1$, $1$) - Upsampling ($2$, $2$) - Convolutional layer: $32 \times 5 \times 5$; stride ($1$, $1$) - Upsampling ($2$, $2$) - Convolutional layer: $2 \times 5 \times 5$; stride ($1$, $1$)
    \item Backward dynamics: $N_z = 10, N_u = 10, N_x = 300 - N_{\text{joint}} = 300 - 16$ units
    \item Planning horizon: $T = 200$
    \item Initial standard deviation for collecting data (SOLAR): $1$ for global and $0.5$ for local training.
\end{itemize}


